\documentclass[11pt]{article}

\usepackage[preprint]{acl}

\usepackage{float}
\usepackage{titlesec} 
\usepackage{fontspec}
\usepackage{caption}
\usepackage{array}
\usepackage{booktabs}
\usepackage{tabularx}
\usepackage{hyperref}

\usepackage{stfloats}

\usepackage{placeins}

\usepackage{fontspec}
\titleformat{\subsubsection}[runin]
  {\normalfont\normalsize\bfseries}{\thesubsubsection}{1em}{}

\newcommand{\wtvt}{\texttt{Wav2vec 2.0}}
\newcommand{\decoar}{\texttt{DeCoAR 2.0}}

\newcommand{\wavlm}{\texttt{WavLM}}
\newcommand{\whisper}{\texttt{Whisper}}

\usepackage{amsmath}

\newcommand{\sonos}{\texttt{Sonos}}
\usepackage{amsmath}
\usepackage{amssymb}
\usepackage{enumitem}
\usepackage{pifont}% http://ctan.org/pkg/pifont
\usepackage{times}
\usepackage{latexsym}

\usepackage[T1]{fontenc}
\usepackage[utf8x]{inputenc}

\usepackage{microtype}

\usepackage{inconsolata}

\usepackage{graphicx}

\title{Identifying and typifying demographic unfairness in phoneme-level embeddings of self-supervised speech recognition models}

\author{Felix Herron\textsuperscript{1,2} \and Solange Rossato\textsuperscript{2} \and Alexandre Allauzen\textsuperscript{1}  \and François Portet\textsuperscript{2} \\
        \textsuperscript{1}MILES Team, LAMSADE, Université Paris Dauphine-PSL, France \\  \textsuperscript{2}GETALP Team, LIG, Université Grenoble Alpes, France\\
        \texttt{felix.herron@univ-grenoble-alpes.fr}}

\begin{document}

\maketitle
\begin{abstract}

Modern automatic speech recognition (ASR) systems have been observed to function better for certain speaker groups (SGs) than others, despite recent gains in overall performance. One potential impediment to progress towards fairer ASR is a more nuanced understanding of the types of modeling errors that speech encoder models make, and in particular the difference between the structure of embeddings for high-performance and low-performance SGs. This paper proposes a framework typifying two types of error that can occur in modeling phonemes in ASR systems: random error/high variance in phoneme embedding, vs systematic error/embedding bias. We find that training phoneme classification probes only on a single, typically disadvantaged SG, sometimes improves performance for that SG, which is evidence for the existence of SG-level bias in phoneme embeddings. On the other hand, we find that speakers and SGs with higher levels of phoneme variance are the same as those with worse phoneme prediction accuracy. We conclude that both types of error are present in phoneme embeddings and both are candidate causes for SG-level unfairness in ASR, though random error is likely a greater hindrance to fairness than systematic error. Furthermore, we find that finetuning encoder models using a fairness-enhancing algorithm (domain enhancing and adversarial training) changes neither the benefits of in-domain phoneme classification probe training, nor measured levels of random embedding error.

% speech encoder models produce phoneme embeddings which are on average both biased and/or higher variance, depending on the SG, for different ages, genders, dialects, ethnicities. However, we find stronger evidence for a link between phoneme prediction error for embeddings with high variance than those with high bias. 

% We describe our methodology, then apply it to a demographically diverse dataset to calculate the sources of error associated with each speaker.

\end{abstract}

\section{Introduction}

Automatic speech recognition (ASR) systems have been repeatedly observed to perform better for certain speaker groups (SGs) than for others \cite{vipperla2008longitudinal,aman2013speech,towards,sonos,meta_fair,gender_perf_gaps_multilingual_pebbles}. Inasmuch there has been ample research in trying to improve ASR fairness \cite{adversarial_doesnt_work_for_unseen,commonvoice_data_based,laura_kmeans,asr_fair_benchmark}; however, resulting fairer models leave ample room for improvement. One potential drawback to existing fairness enhancing studies is that they tend to treat ASR systems like black boxes - it is hard to design an intervention to reduce demographic unfairness without better understanding the mechanisms involved in engendering unfairness to begin with. This paper provides a framework for understanding two sources of modeling error with respect to unfair SG-level treatment (see Figure \ref{fig:error_typology}): \textbf{embedding bias/systematic error}, in which phonemes are modeled around different modes depending on SG; vs \textbf{unequal variance/random error}, in which phonemes are modeled around the same modes but with more noise for one SG than another.

We perform probing experiments on latent embeddings of  state of the art attention-based speech encoder models, in particular on ASR-finetuned self-supervised speech processing models (S3Ms), to evaluate the bias and variance of phoneme embeddings across SGs. First, we evaluate whether phoneme recognition (PR) probes trained on a single SG perform better on unseen speakers from that same SG and worse for unseen speakers from unseen SGs. We find statistically significant improvement for in-domain training for some but not all SGs, at some but not all layers, for some but not all phonemes, thus indicating the existence of \textbf{embedding bias} in phoneme embedding at those layers for those models. However, we find that regardless of training data, the same SGs almost always perform best or worst respectively on PR, thus indicating the prevalence of higher levels of \textbf{embedding variance} for some SGs.

Second, we directly calculate the  variance in phoneme embeddings using a k-nearest neighbors (KNN) distance heuristic. We find that the same SGs which experience worse PR performance also have higher variance in PR, and there is statistically significant correlation between the two metrics for every SG and every speech encoder model.

Finally, we investigate the difference between how phonemes are embedded for different SGs when we finetune for ASR using the fairness enhancing algorithm Domain Enhancing/Adversarial Training (DET/DAT). We find that it changes neither the relative KNN distance between phonemes, nor the benefit achieved from in-domain PR training. Thus, we conclude that to the extent that KNN distance is a useful indicator of precise phoneme modeling, that DET/DAT, as constituted in our study, is not useful at rendering the phoneme embedding space fairer.

% embeddings on normal pretrained models vs models that have been pretrained using domain adversarial training (DAT) to be blind to 

%. To measure by measuring the extent to which phoneme embeddings can be used to predict SG

% Classical phonetics provides evidence for systematically variable embeddings based on measurements such as vowels projected onto the F1-F2 space \cite{bing_youtube_dialect}; however, to what extent those differences are retained in the layers which maximize ASR performance in deep ASM's had not yet been studied.

\section{Related work in interpretability of speech processing systems}

\citet{toyota_layerwise_interpretability,toyota_layerwise_interpretability_2} show that certain types of information inherent to or contained within an utterance, such as speaker identity or word meaning, are maximized at different model layers of S3Ms. \citet{pastorExplainingSpeechClassification2024} use explainable AI methods to show the impact of linguistic and paralinguistic features on transcription in \wtvt{} models. \citet{masson_phoneme_similarity_tts} find that artificially generated non-native speech is represented in a similar manner to real non-native speech in ASR systems. This suggests that ASR systems model non-native speech in a systematically different and reproducible manner, and anecdotally supports the \textbf{embedding bias hypothesis} (see Figure \ref{fig:error_typology}). \citet{towards,dutch_quantifying} find many of the same phonemes are hardest for ASR systems to understand regardless of SG or ASR architecture. 

\citet{orthogonality_first,crv_pca} show that phoneme information and speaker information are modeled in orthogonal subspaces of S3M embeddings. This would suggest that lack of demographic parity in ASR is unlikely due to speaker-specific phoneme modeling differences, providing anecdotal support of the \textbf{higher variance hypothesis}. \citet{layerwise_fairness_interspeech_todo_arxiv} shows that layers of pretrained self-supervised speech processing models (S3Ms) that are best capable of modeling ASR are also the least fair, and that this effect is not attenuated after ASR finetuning. Furthermore, they show that even S3Ms pretrained on non-English corpora display the same biases as S3Ms pretrained only on English. These findings motivate the notion that certain speakers are inherently harder to model, regardless of training data and model architecture. Finally, \citet{herron_tsd} shows that pretrained S3Ms model certain SGs in a differentiable manner on an utterance scale, though to a lesser extent after ASR finetuning; \citet{accent_probing_vs_cca} shows similar results for accent encoding. 

Other studies have delved into ASR errors due to hallucination in decoding \cite{openai_4o_voice,asr_hallucination,asr_hallucinate_non_speech,whisper_hallucination}. Our study doesn't touch on this - we consider only speech \textit{encoders}, and perform PR using a single ``decoder'' linear layer to map from embedding space  to phoneme space. Furthermore, we simplify the task by avoiding the challenges of sequence-to-sequence mapping, instead focusing on single isolated phoneme embeddings. However, it is important to note that the two types of encoding error which we study are not the only potential sources of error in ASR.

\section{Motivation}

\begin{figure}[t]
  \centering
  \includegraphics[width=\linewidth]{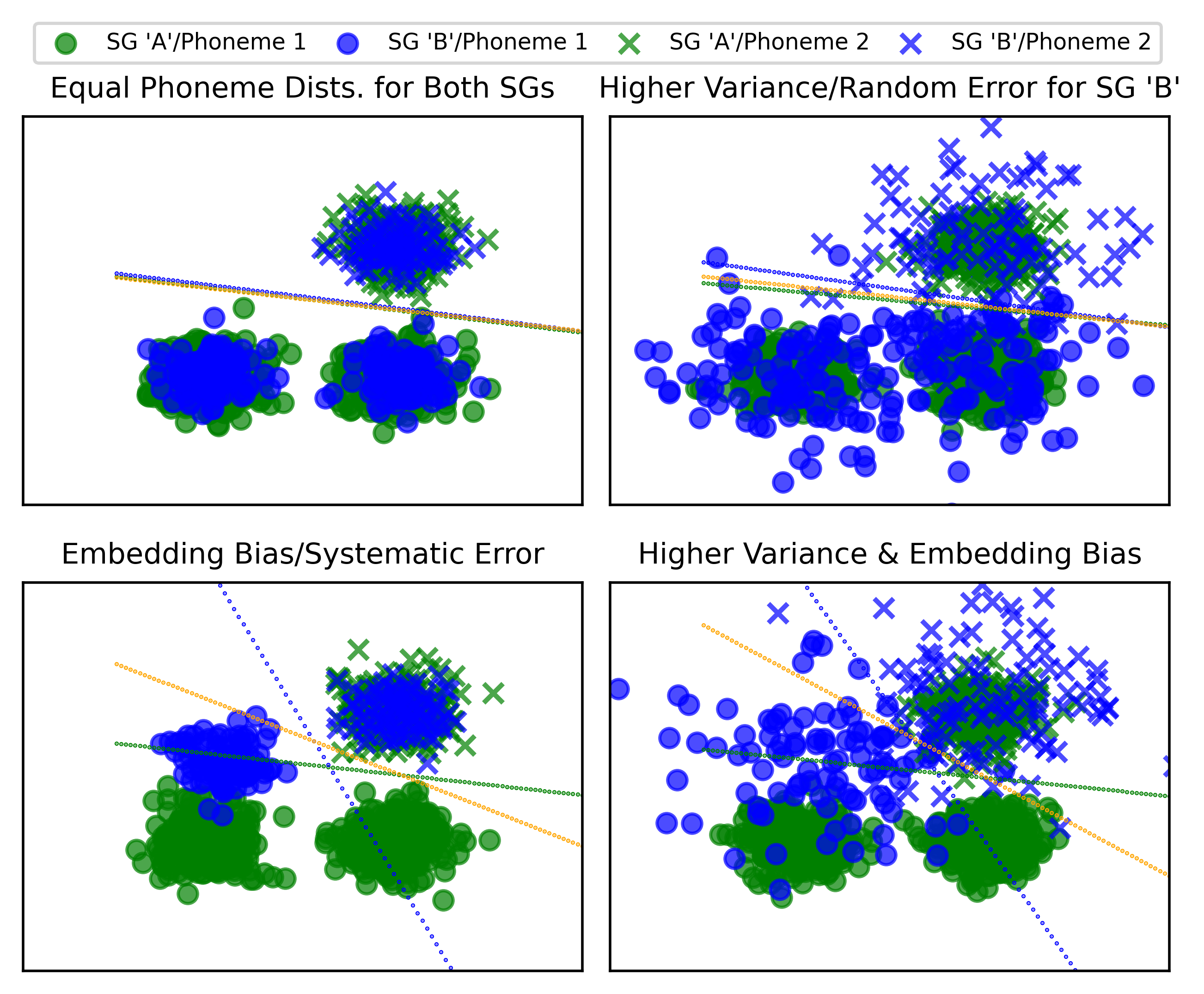}
  \caption{Toy visualization of high variance vs embedding bias in by-SG modeling of phonemes, along with corresponding linear separators trained on the SG of the corresponding cover (orange trained on both SGs). If, compared to an equal-error world (upper left), the S3M embeds phonemes from SG 'B' with \textbf{higher variance} than SG 'A' (upper right), the learned ASR head won't necessarily learn a different transformation (thought it might due to the training data being less noisy) but it will return higher error rate for SG 'B'. However, if phoneme 1 is modeled about different modes depending on SG, representing \textbf{embedding bias} (lower left), this will bias the PR head towards the more heavily represented SG in the training set.}
  \label{fig:error_typology}
\end{figure}

The objective of this paper is to examine the differences between how S3Ms model phonemes of various SGs. In an ideal world for ASR, each phoneme embedding should unambiguously be represented by a vector that can be linearly mapped to its corresponding phoneme label, regardless of speaker or SG. However, sometimes it is ambiguous which phoneme an embedding is representing, which is one reason why ASR systems can make mistakes. If we consider a distribution over embeddings for a group phonemes for a given SG, we identify two mechanisms by which they might be misclassified. This analysis is based on a classical error dichotomy in classification error of random vs systematic error \cite{error_analysis_stats}, as illustrated in Figure \ref{fig:error_typology} on toy data. \textbf{1) embedding bias/systematic error}, in which a phoneme is embedded about a different mode than the linear classifier has been trained on, due to particularities of a SG; or \textbf{2) high variance/random error}, in which a phoneme is embedded about a correct mode in embedding space, but with greater noise for one SG than another. Or \textbf{3) both}: in all likelihood, a mixture of both types of errors occurs at some level in phoneme embedding. We seek to measure whether one error type better explains unfairness for some SGs than for others.

%\ref{fig:error_typology} illustrates a simplified version of this framework for toy data.

%There is likely a mixture of both error types occurring for all SG unfairness - 

% In simpler terms, we are trying to test whether SG-level unfairness in tems is more due to imprecise modeling of certain phonemes for certain SGs, or due to 

\section{Experimental setup}
\subsection{Data preparation}

We use the Sonos Voice Control Bias Assessment Dataset for our analyses \cite{sonos}. Its recordings are clean and of similar fidelity across speakers, thus reducing bias deriving from varied access to high-quality recording equipment and environments \cite{noise_means_bad_asr}. Furthermore, \sonos{} provides SG-level metadata for speakers' gender, age, dialect, and ethnicity.

We start by aggregating some SGs to avoid accidentally biasing our results due to imbalance in dataset construction \cite{herron_speakable}. To wit, we aggregate all native speakers of different mainland USA dialects (of which there are 6 in \sonos{}) into a "Native" category; we do the same for middle-aged adults (of which there are several arbitrarily delineated categories) into an "adult" category. Our aggregations are grounded in 1) a lack of expected modeling difference based on theoretical linguistic theory of how voices age \cite{old_voice_change}, and 2) empirical results of previous studies which have analyzed this dataset and found little difference between adult age groups or Native speakers of American English \cite{sonos,layerwise_fairness_interspeech_todo_arxiv}. We also remove the oldest age category in \sonos{}, 55-100; it is a potentially exceptionally heterogeneous category, adding noise to the "adult" SG.

We are thus left with 951 speakers distributed over gender (male, female), dialect (Native, Latino, Asian), age (children, adults), and ethnicity (Caucasian, African-American)\footnote{Ethnicity labels are only available for a small subset of the dataset (98 speakers).}.

When defining SG partitions (i.e. defining the speakers who fall under ``men'' vs ``women''), it is important to consider the multifaceted identities of individual speakers \cite{herron_speakable}. \sonos{} provides SG labels for four demographic variables - we take advantage of these to isolate \textit{single} demographic variables and removing influence of other potentially confounding variables. For example, if most men in the dataset are African American and most women are Caucasian, when measuring the performance for men vs women we are likely to inadvertently measure the effect of ethnicity in addition to gender \cite{meta_fair}. To counteract this, we exclude speakers from non-mode SGs (with respect to the dataset's construction) SGs: when testing the effect of gender, we include neither children, non-native speakers, or speakers with labeled ethnicity\footnote{By using only speakers with no ethnicity tag, at least we avoid measurably oversampling one ethnicity over another. However, this is a potential source of bias in our analysis, given that the well-established impact ethnicity has on ASR \cite{habitual_be}.}; when testing the effect of age, we include no non-native speakers and speakers with ethnicity labeled; when testing the effect of dialect, we include no children.

Retaining only mode speakers has the advantage of avoiding certain sources of dataset-driven bias, but adds a fresh layer of experimental bias by not taking uncommon SG combinations into account. By ignoring non-native children, for example, we are unable to measure the potentially confounding effect of the intersectionality of those SGs together \cite{intersectionality_ml_survey,towards}.

% This result is not terribly surprising, given the multitudinous dimensions along which SG information has been found to be modeled in S3Ms .

\subsection{Backbone Encoder Models}

This paper studies speech encoder models, and we focus in particular on self-supervised speech processing models (S3Ms) \cite{wav2vec2}. S3Ms use only audio during pretraining, isolating them from potential bias that comes from finetuning for ASR. Furthermore, pretrained S3Ms have been shown to model phonetic rather than semantic aspects of speech \cite{phonetics_over_semantics_in_s3m}. This insulates them from bias related to semantics, which facilitates our error analysis. S3Ms are also useful starting points for transfer learning and can be finetuned using little labeled training data to solve diverse downstream tasks, from ASR to SID \cite{wav2vec2}. Though not an S3M, we also study the \whisper{} encoder, which was trained end-to-end on ASR. \whisper{} achieves state of the art ASR performance, so it is informative to understand its behavior \cite{whisper}.

The encoder models we study include \wavlm{}-base-plus (WavLM-base+), \wavlm{}-large, \wtvt{}-large-ls (W2V2-lg-ls), \wtvt{}-large-xlsr-53 (XLS-R), \decoar{} (DeCoAR2), and \whisper{}-medium (Whisper-med) - one for each of the three families of S3Ms according to \citet{types_of_s3ms}. We choose \wavlm{} models \cite{wavlm} as they have been empirically shown to create the best universal speech embeddings for English of any S3M based on the SUPERB benchmark \cite{superb}. We compare between the base and large ($\approx 100M$ and $\approx 300M$ parameters respectively) model sizes, both of which were trained on the 60k hour LibriVox corpus. We also include two \wtvt{} models \cite{wav2vec2}, as it remains the most popular S3M according to Huggingface downloads. Furthermore, we can compare the effect of pretraining on a small, constrained, and regular corpus like LibriSpeech in W2V2-lg-ls vs a more varied, multilingual corpus in XLS-R \cite{w2v2-xlsr}. Finally, we consider the encoder module of \whisper{}-medium \cite{whisper}, which has the same number of parameters as the large S3Ms. The pretraining data for \whisper{} are not public which makes it more challenging to compare with its peers. We used Speechbrain to extract phoneme embeddings and train our PR probes \cite{speechbrain}. See Appendix Section \ref{sec:embedding_extraction} for further technical details about phoneme embedding extraction.

We finetune each pretrained S3M for ASR using the same data and training loop to ensure optimal comparability. All our adaptations are trained based on Speechbrain recipes using a subset of 1500 speakers sampled from CommonVoice 16 \cite{common_voice}. We finetune using using CTC loss and a 3-layer MLP decoder. We first freeze the S3M and train the decoder until convergence, then unfreeze the S3M and train for 30k steps.

\subsubsection{Domain Enhancing/Adversarial Finetuning}

One common technique for improving fairness in speech processing is domain enhancing or adversarial training (DET/DAT), in which the model is forced to maximize or minimize some type of information at specific layers \cite{adversarial_and_enhancing,mtl_dat_comp,dat_english_accents}. Typically, this involves learning a classifier for speaker attributes (e.g. accent or speaker ID), and using its gradient to force the backbone encoder model to create embeddings that maximize speaker information in middle layers and minimize it in final layers (using a gradient reversal layer for adversarial training). This is typically applied as a multi-task objective during ASR finetuning (CTC + DET/DAT).

%DET/DAT cannot be applied in isolation as the model would be corrupted without an objective forcing it to continue modeling speech properly (such as self-supervised in pretraining or ASR in finetuning) \cite{forgetting_ssl}.

Empirical results using CTC + DET/DAT have not been exemplary \cite{dat_english_accents,survey_on_accents_det_dat,adversarial_doesnt_work_for_unseen,domain_adversarial,fairspeech_bad_methodology_asr}. Recent work has shown that this could be due to models being already irreparably biased during pretraining, and that fairness interventions ought to come \textit{before} ASR finetuning \cite{layerwise_fairness_interspeech_todo_arxiv}, as well as that ASR finetuning by itself tends to cause encoder models to discard most speaker information by the final layer \cite{herron_tsd}. However, we are interested to see the effects of DET/DAT on the phoneme-embedding level. We follow \citet{adversarial_and_enhancing}, learning enhancing and adversarial speaker ID classifiers on layers 5 and 10 for base models, and 10 and 21 for large models respectively. One advantage of using speaker ID as a classification target is that it does not depend on potentially fuzzy SG labels. Following \citet{adversarial_doesnt_work_for_unseen}, we first warm up the model to conversion on both classifiers before unfreezing its weights. Following \citet{herron_tsd}, we use an x-vector as our adversarial classifier to avoid learning a too narrow classifier and retaining speaker information in subspaces orthogonal to the linear classifier \cite{xvectors}. Following \citet{toyota_layerwise_interpretability}, we re-initialized the final two layers of \wtvt{} models, as they are overfit to their initial contrastive objective and do not provide good starting points for other speech modeling tasks (such as ASR).

% For continual pretraining, we adopt an algorithm based on the second iteration of HuBERT \cite{hubert}. We learn a 500-centroid k-means clustering of frame embeddings at layer 6 or 12 for base and large models respectively, then train a linear layer after the final transformer layer to predict the centroid identity of each masked frame. 

\section{Methodology}

We perform two types of experiments to measure both the variance and bias in phoneme embeddings for each SG in the \sonos{} corpus.

\subsection{Indirectly measuring embedding variance vs bias via PR probes}
\label{sec:indirect_explanation}

For our first test, we train phoneme classification probes on varying subsets of the extracted (and frozen) phoneme embeddings using a \textbf{linear} probe. We chose the simplest possible architecture in order to maximize the likelihood that bias uncovered during probe training is reflects bias in the underlying S3M, rather than more complex bias learned during training of the probe. (The one obvious downside of this choice is that real life ASR use much deeper decoders than just linear layers). For each demographic variable (gender, age, dialect, ethnicity), we have \textbf{two probe training settings}:

\begin{enumerate}
    \item Train the linear probe using data from only a single SG (e.g. men). We use the same number of speakers no matter which SG\footnote{Previous studies have shown the number of speakers to be a critical factor in generalizability of ASR systems \cite{commonvoice_data_based,more_speakers_in_training_set,ryan_data_selection}. Even though this is just a probe, we are fastidious in equally balancing numbers of speakers to avoid conclusions based on random dataset imbalance.} - (e.g. the same number of men in the men-only as the number of women in the women-only).
    \item Train the linear probe using data from a balanced number of speakers from each SG (i.e. the same number of men as women). Critically, we use the same number of speakers overall as in the previous setting to ensure direct comparability. \label{sec:probe_train_setting_balanced}
\end{enumerate}

\noindent We then measure the macro F1 accuracy for each classifier in predicting each phoneme for each SG under each setting, at each layer of each encoder model. We replicate each experiment five times using different speakers for each replication\footnote{Apart from the SG (e.g. men) with minimum number of speakers for its corresponding demographic variable (e.g. gender), which defines the number of speakers used for all SGs for said demographic variable (\sonos{} contains fewer men than women, so all PR probes will be trained with the number of speakers as there are men).}. Such training-set balancing is a common strategy in fairness research \cite{hend_gender,solange_balanced}, though it is often used in training or finetuning entire networks. Our intervention has the potential to impact far fewer parameters (only those of the phoneme probe), so we are likely to see more consistent results than those observed in aforementioned studies.

\subsection{Random error as KNN distance}

We directly measure the random error of phoneme embeddings for each speaker of each SG (Figure \ref{fig:error_typology} upper right) by calculating the KNN distance between phoneme embeddings for individual speakers to determine how closely clustered they are. See Appendix Section \ref{sec:knn_motivation} for an explanation as to why KNN distance is an appropriate metric for measuring random error in phoneme embeddings.

Our protocol is as follows: fixing a speaker $s$ and layer $M_\ell$, we first normalize over all extracted phoneme embeddings to ensure unit-mean/standard deviation. Then, we reduce dimensionality by retaining dimensions which account for 95\% of all variance between embeddings for all phonemes, using principal component analysis (PCA). This greatly reduces the size of embeddings and potentially winnows out noise along axes tangential to phoneme identity. Then, for each phoneme $p$, we calculate the KNN distance as the squared L2 distance from each embedded sample of $p$ (i.e. $p^{s, M_{\ell}}_i$ for $i \in [1..30]$) to its k-nearest neighbors of the same phoneme (i.e. $p^{s, M_{\ell}}_{\{j,k,l\}}$ for $i \notin \{j,k,l\}$). We use the same number of samples N=30 and k=3 for each speaker/phoneme to ensure comparability between phonemes, speakers, and layers. We then aggregate over all speakers for a SG to determine a distribution of KNN distance over each SG. If $M_{\ell}$ produces embeddings with higher KNN distance for one SG than another, that is evidence of higher random error in the way $M_{\ell}$ models $p$ for SG.

\section{Phoneme recognition probing results}

\begin{figure}[t]
  \centering
  \includegraphics[width=\linewidth]{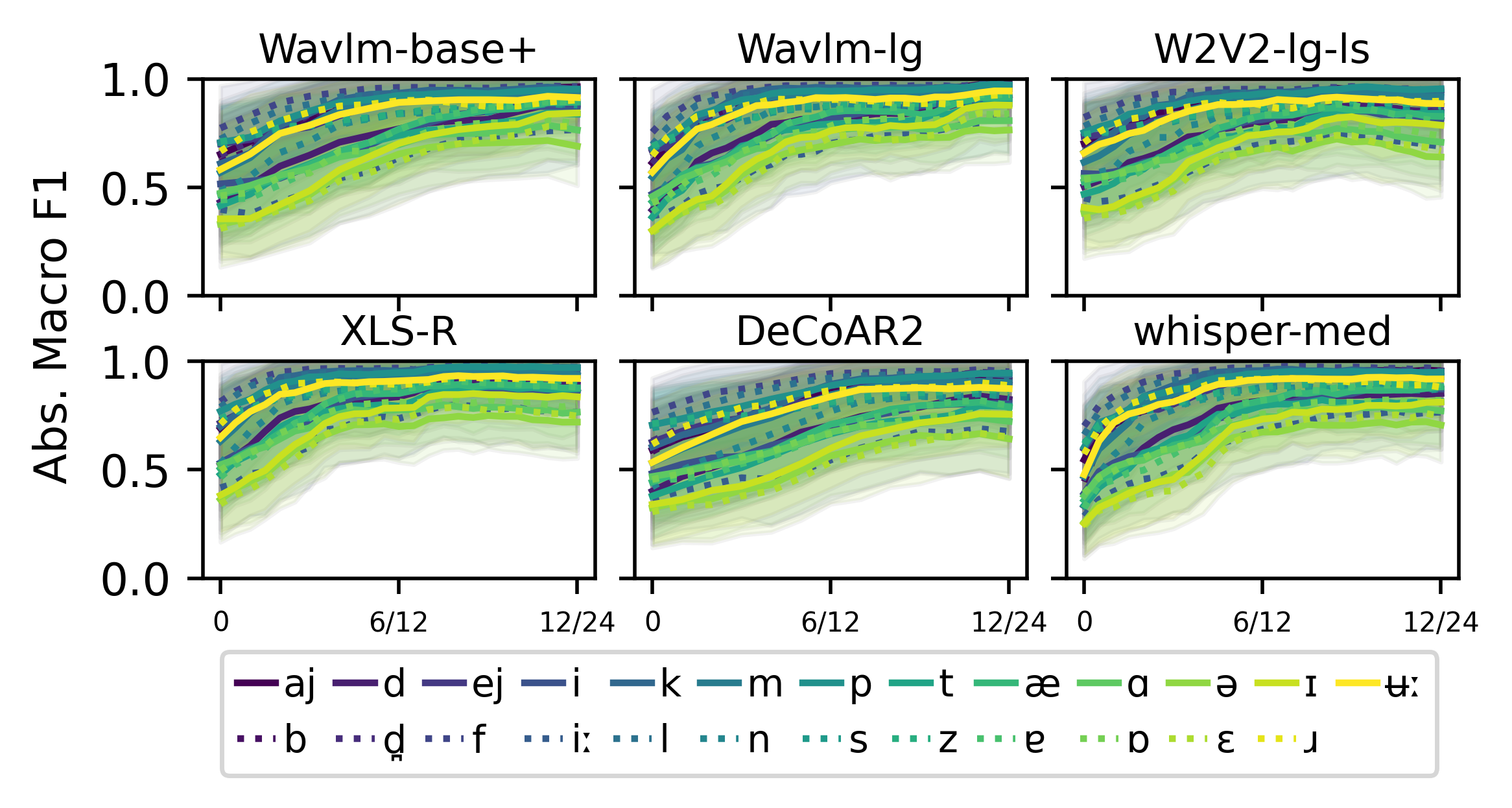}
  \caption{Absolute F1 score at every layer for each phoneme of each ASR-finetuned encoder model.}
  \label{fig:phoneme_rec_overall}
\end{figure}

\begingroup
\setlength{\tabcolsep}{5pt} % Default value: 6pt
\renewcommand{\arraystretch}{0.5} % Default value: 1

\begin{table}[t!]
    \caption{Overall F1 phoneme classification rate for best overall layer of each ASR-finetuned encoder model, for PR probes trained on balanced data for each demographic variable (train setting \ref{sec:probe_train_setting_balanced}). Gap for each demographic variable is the percent difference between the worst and best performing SGs \cite{amazon_kmeans}.}
    \scriptsize
  \centering
  \centerline{
  \addtolength{\tabcolsep}{-0.3em}
    %%%insert tabular here

\begin{tabular}{l@{\hspace{-2.2em}}rrrrrr}
% \midrule
& Wavlm-base+ & Wavlm-lg & W2V2-lg & XLS-R & DeCoAR2 & whisper-med \\\midrule
Macro F1 $\uparrow$ &0.88 & 0.91 & 0.87 & 0.90 & 0.84 & 0.88 \\
Gender gap $\downarrow$ & 0.08 & 0.64 & 0.21 & 0.21 & 0.86 & 0.03 \\
Age gap $\downarrow$ & 3.85 & 2.74 & 4.18 & 3.02 & 3.95 & 3.51 \\
Dialect gap $\downarrow$ &6.26 & 4.48 & 7.70 & 6.23 & 7.28 & 6.33 \\
Ethnicity gap $\downarrow$ & 2.39 & 2.43 & 3.46 & 2.34 & 3.01 & 1.90 \\
% \bottomrule
\end{tabular}
    }
    % \caption*{Greetings}

     \label{tab:phoneme_rec}
\end{table}
\endgroup

While some models have higher overall performance (\wavlm{}-large is the best, see Table \ref{tab:phoneme_rec} row 1), and some are slightly fairer than others (\hyperref[tab:phoneme_rec]{rows 2-5}), we find that all the models behave roughly equivalently with respect to our in-domain probing experiments. Furthermore as Figure \ref{fig:phoneme_rec_overall} shows, some phonemes are harder to identify overall and all models struggle with the same phonemes. We also replicate these experiments on pretrained S3Ms and find similar results (see Appendix Section \ref{sec:pretrained_PR}) - thus we can rule out bias acquired during ASR finetuning as responsible for our findings.

We also note an overall lack of evidence supporting embedding bias in later layers of finetuned S3Ms. When compared to pretrained models (see Appendix Section \ref{sec:pretrained_PR}), finetuning for ASR seems to reduce any bias that might have persisted. This is not terribly surprising - \citet{herron_tsd} shows that SG information is dropped during ASR finetuning, thus eliminating a source of bias.

\begin{figure}[t]
  \centering
  \includegraphics[width=\linewidth]{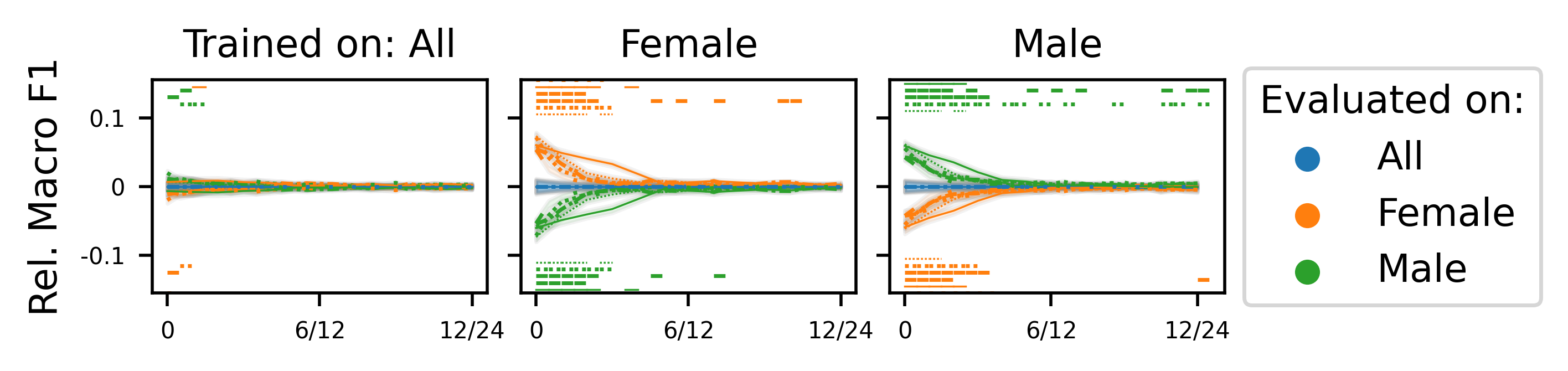}\hfill
  \includegraphics[width=\linewidth]{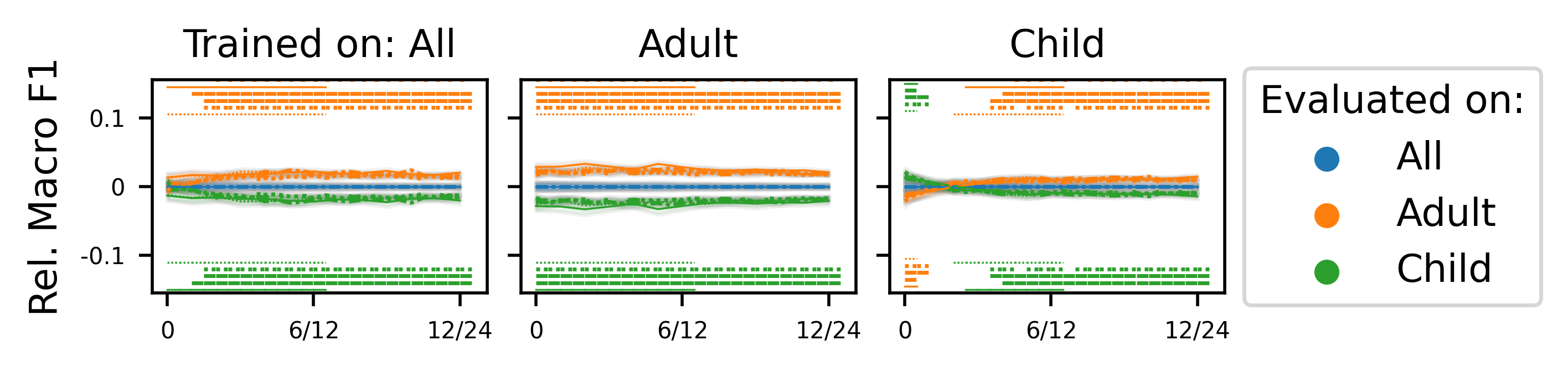}\hfill
  \includegraphics[width=\linewidth]{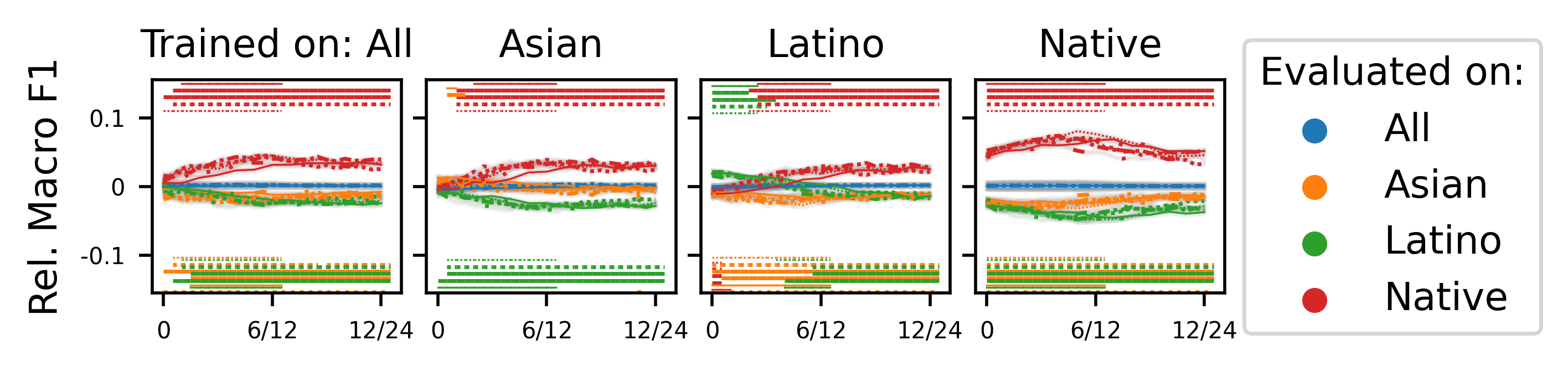}\hfill
  \includegraphics[width=\linewidth]{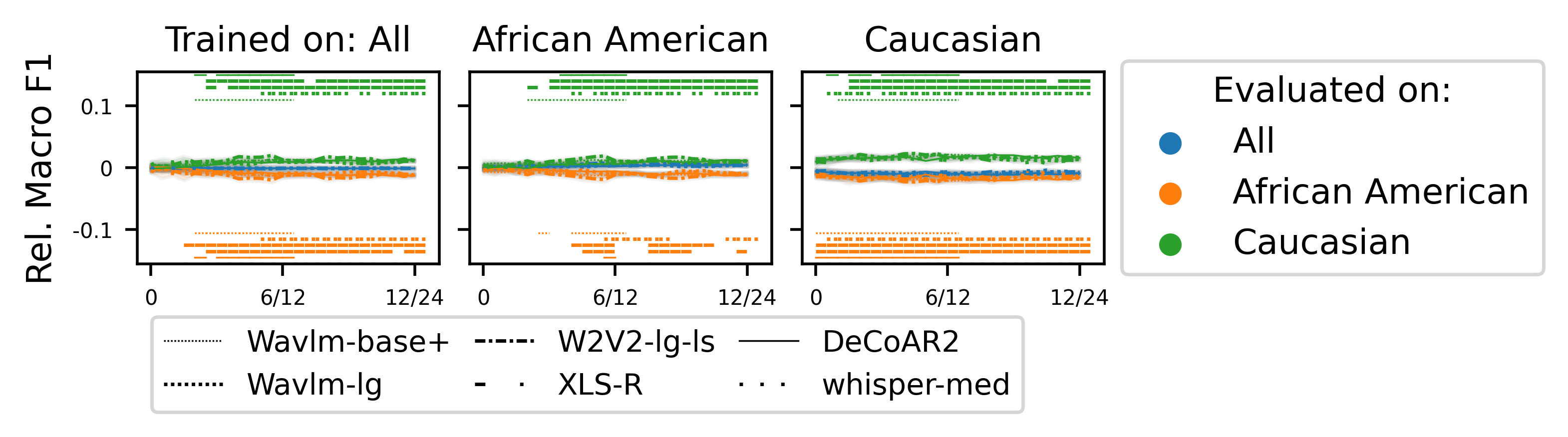}
  \caption{Macro F1 phoneme classification scores, \textbf{relative to the macro average over all SGs} (e.g. men and women) for the corresponding demographic variable (e.g. gender) for ASR-finetuned S3Ms. Values $>0$ indicate that SG has a  better-than-average macro F1 (e.g. top left: when trained on all data, females have above-average macro F1 performance in layer 0). Horizontal lines on top and bottom of each figure denote statistical significance (for $p<0.05$) for relative F1 $</> 0$ respectively on 1-sided 1-sample t-test.}
  \label{fig:by_training_finetune}
\end{figure}

\begin{figure}[t]
  \centering
  \includegraphics[width=\linewidth]{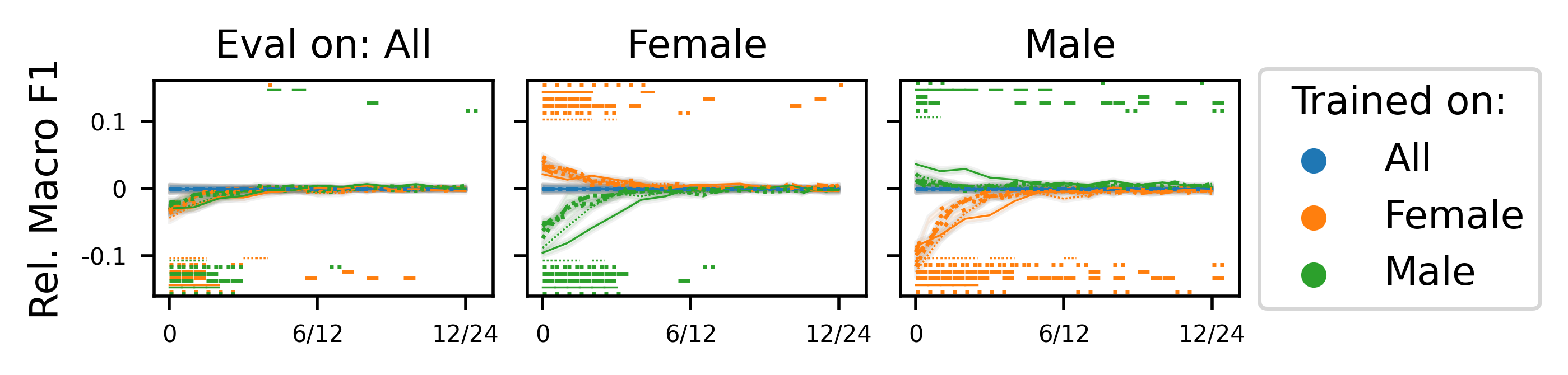}\hfill
  \includegraphics[width=\linewidth]{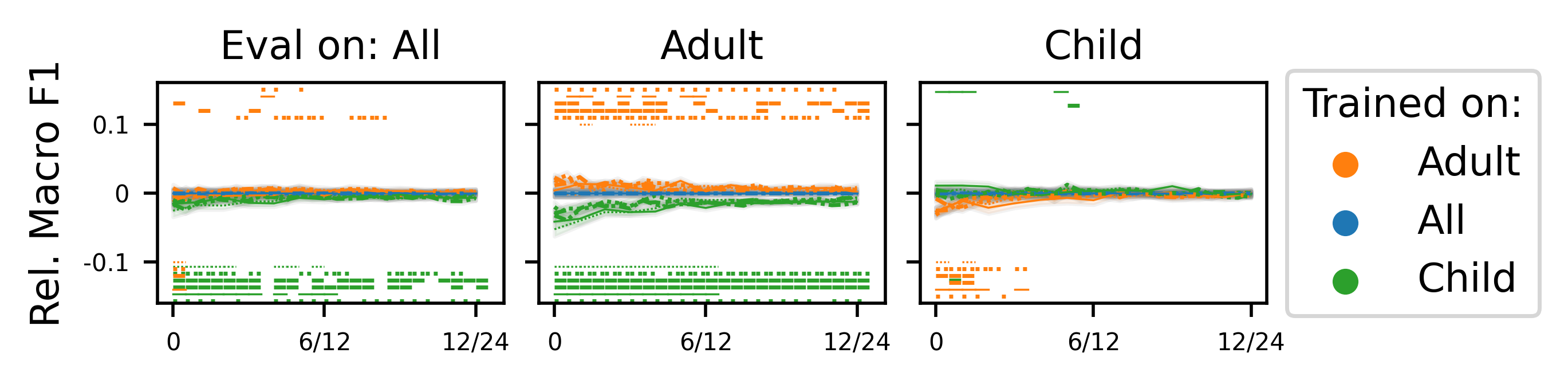}\hfill
  \includegraphics[width=\linewidth]{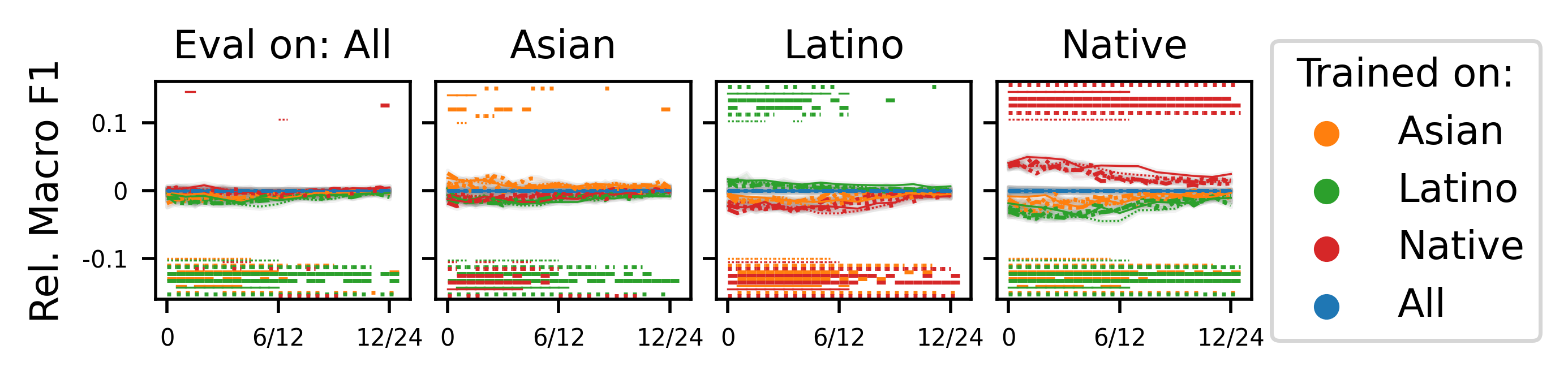}\hfill
  \includegraphics[width=\linewidth]{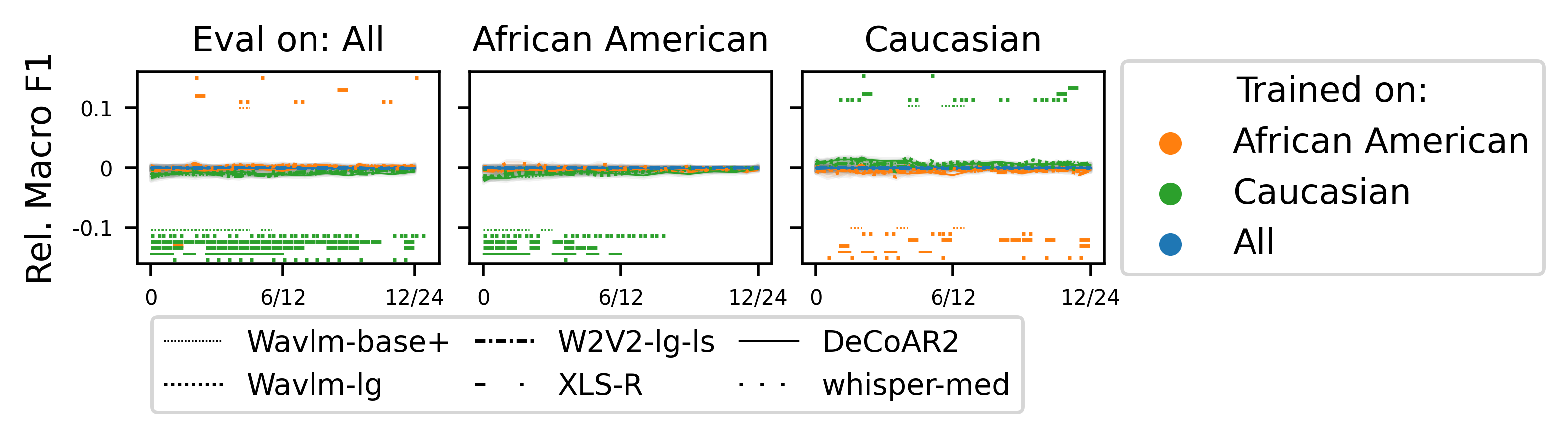}
  \caption{Macro F1 phoneme classification scores,\textbf{ relative to probe training on a balanced dataset} for SGs from the corresponding demographic variable for ASR-finetuned S3Ms. Values $>0$ indicate that training on that SG results in better performance than for a balanced dataset (e.g. top left: when phoneme classifiers are trained on only men, men have a higher PR than for probe training on balanced data). Horizontal lines on top and bottom of each figure denote statistical significance (for $p<0.05$) for relative F1 $</> 0$ respectively on 1-sided 1-sample t-test.}
  \label{fig:by_eval_finetune}
\end{figure}

\subsection{Analysis of in-domain training overall}

Figures \ref{fig:by_training_finetune} and \ref{fig:by_eval_finetune} show relative macro F1 scores for S3Ms (and \whisper{}) finetuned for ASR. The first figure shows the effect on all SGs when varying training data; the latter shows the effectiveness of training on each SG when varying the test set. We compute 1-sample 1-sided t-tests for each model, layer, and SG, to compare it with the baseline - for Figure \ref{fig:by_training_finetune}, the baseline is average F1 over all SGs; for Figure \ref{fig:by_eval_finetune}, the baseline is by-SG performance for probes trained on balanced training data.

\subsubsection{Does SG-specific probe training change relative performance among SGs?} 

When examining the left-most columns of Figure \ref{fig:by_training_finetune}, for probes trained on a balanced group of all SGs, our results overall are in line with previous studies analyzing \sonos{} \cite{sonos,layerwise_fairness_interspeech_todo_arxiv} - native speakers, adults, and Caucasians have the best performance for all models, while there is no significant difference for gender.

Furthermore, when regarding the other columns (phoneme probes trained on speakers from a single SG) we see that in-domain training has little effect on the pecking order of relative PR performance. The greatest divergence from balanced training comes in the first several layers of each model, where the polarity sometimes flips (for age and dialect, for example). However, those layers have poor overall PR performance (see Figure \ref{fig:phoneme_rec_overall}). For later layers with better overall PR, the same SGs get the best PR performance \textbf{regardless of which SG is used in training the PR probes}. This result sets a low ceiling for the amount of embedding bias/systematic error in phoneme embeddings for any of the models we tested. It also corroborates the finding in \citet{orthogonality_first} that speaker and phoneme information are modeled orthogonally - if there were more mixture of the two, then we would expect performance improvements for in-domain phoneme probe training.

\subsubsection{Does in-domain training help individual SGs?}

Figure \ref{fig:by_eval_finetune} shows the effect on individual SGs when training phoneme probes on speakers from a single SG. For overall performance, we observe that training on a balanced dataset is almost always best. For gender, we observe that in-domain probe training is beneficial for both females and males in early layers but that they experience less boost in later layers (though still statistically significantly so for some S3Ms). At the same time, the out-of-domain gender experiences degraded performance.

For age, we observe that training on only children benefits no one, not even children, while training on only adults is beneficial for adults for most layers of most models. We interpret this as emblematic of greater variance in children's phoneme embeddings - in-domain training on noisier data results in a worse calibrated classifier, and testing on children's data is inherently harder due to it being noisier. If there were significant embedding bias, we would observe increased performance for children on probes trained only on children.

For dialect, we observe a clearer example of the potential benefit of in-domain training, albeit only barely (and not significantly for all models), likely indicating embedding bias. Asian, Latino, and Native speakers (row 3, columns 2, 3, and 4 respectively) achieve their best performance at many layers, though most substantially early layers, when trained on only similar speakers. For ethnicity, we observe minimal advantage gained from in-domain training for Caucasian speakers but none for African-American speakers.

It is important to note that these are macro F1 scores over all phonemes, and might not highlight bias in individual phonemes if their effect is small and they are few in number. Appendix Section \ref{sec:by_phoneme_analysis_probe_training} shows that some phonemes benefit more than others from in-domain probe training, indicating that \textit{there is} SG-level embedding bias for some phonemes, just not all.

\section{KNN distance results}

\begin{figure}[t]
  \centering
  \includegraphics[width=\linewidth]{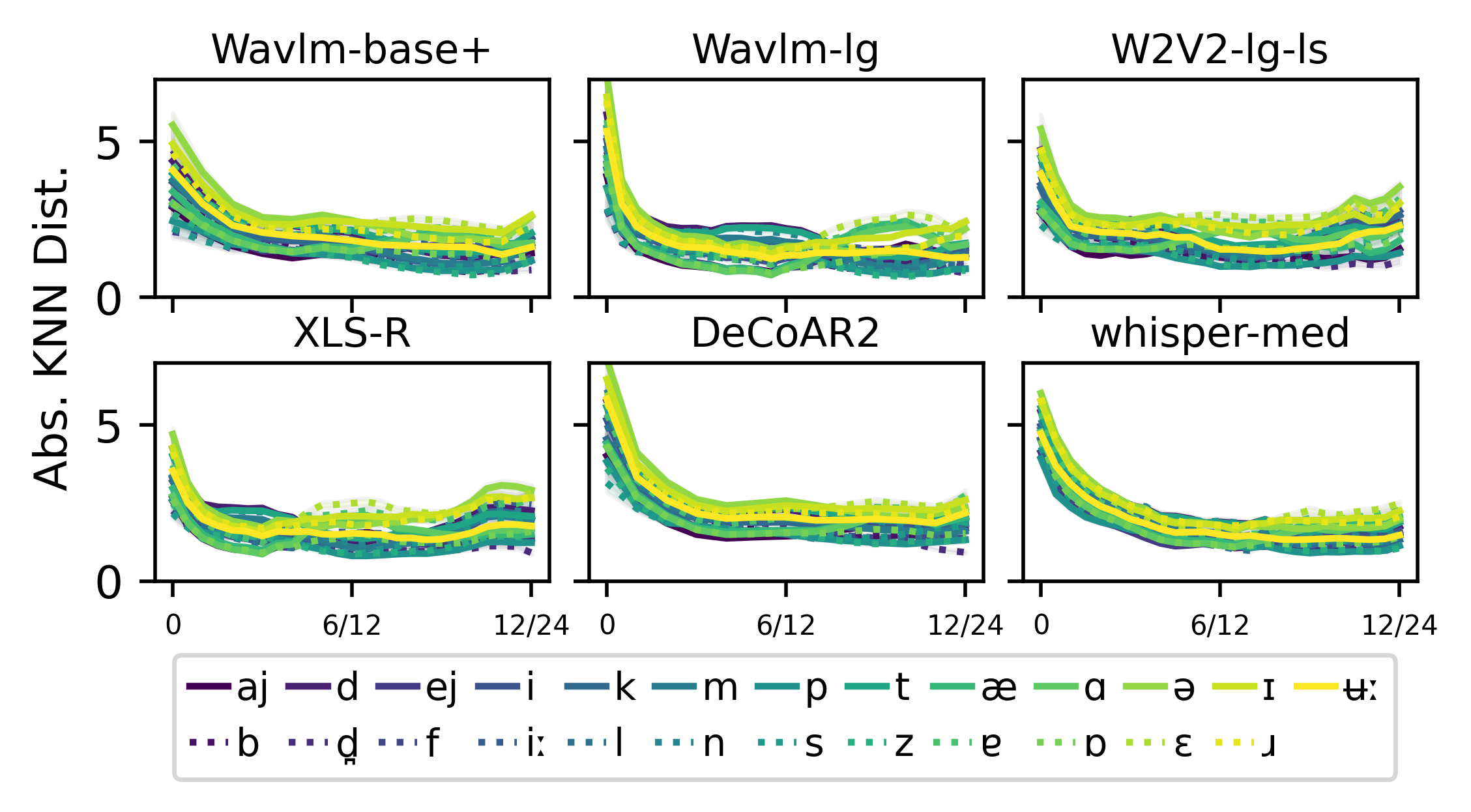}\hfill
  \caption{Absolute KNN distance for each phoneme, averaged over all speakers. Higher values for KNN distance represent greater variance in phoneme embedding.}
  \label{fig:knn_distance_absolute}
\end{figure}

\begin{figure}[t]
  \centering
  \includegraphics[width=\linewidth]{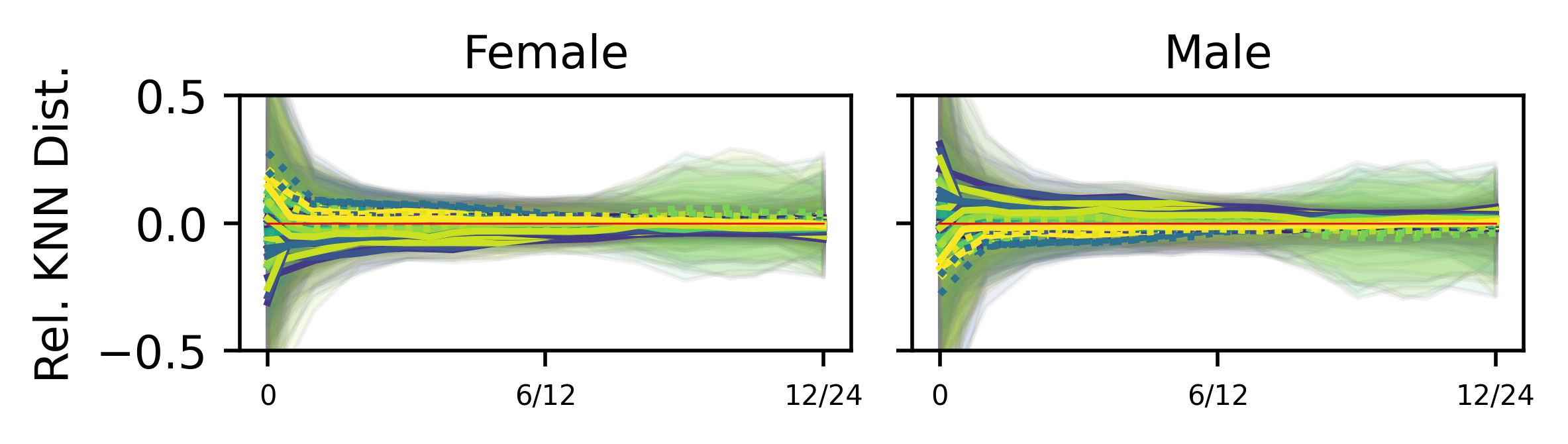}\hfill
\includegraphics[width=\linewidth]{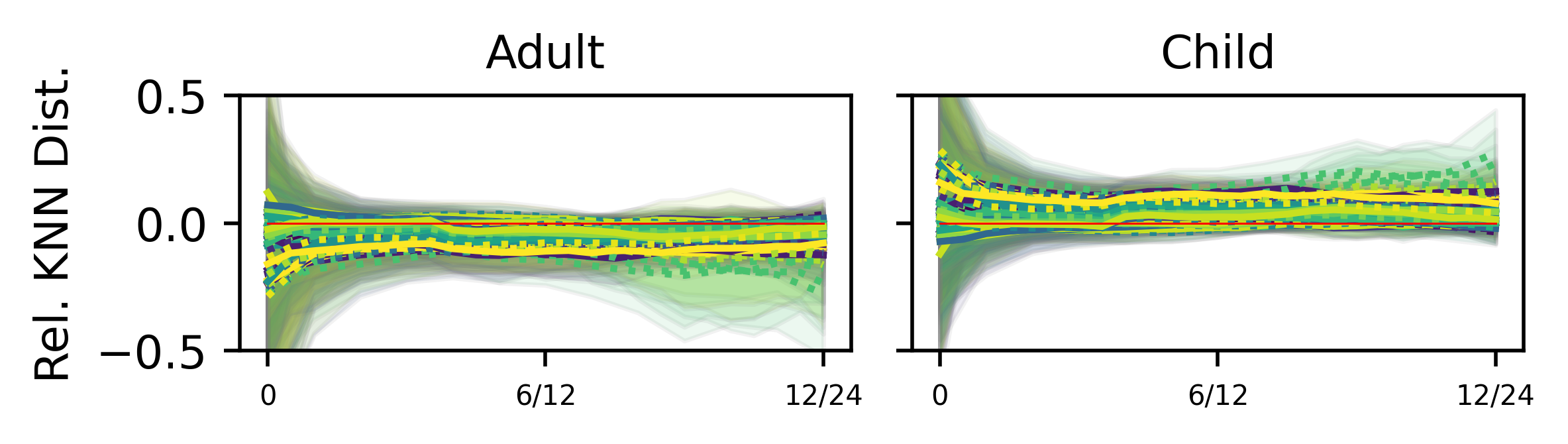}\hfill
\includegraphics[width=\linewidth]{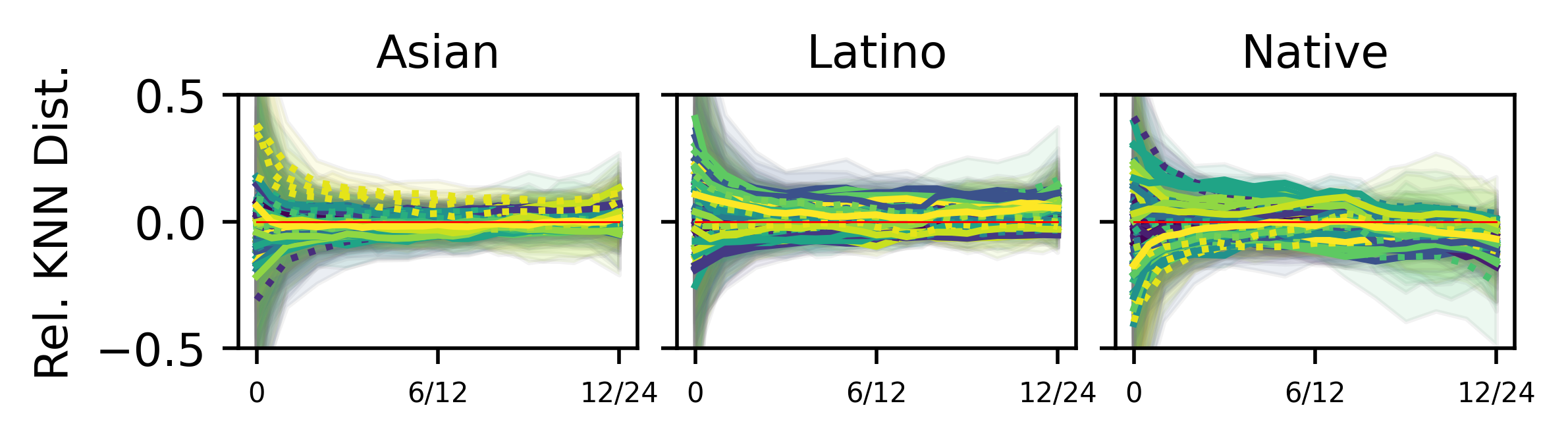}\hfill
\includegraphics[width=\linewidth]{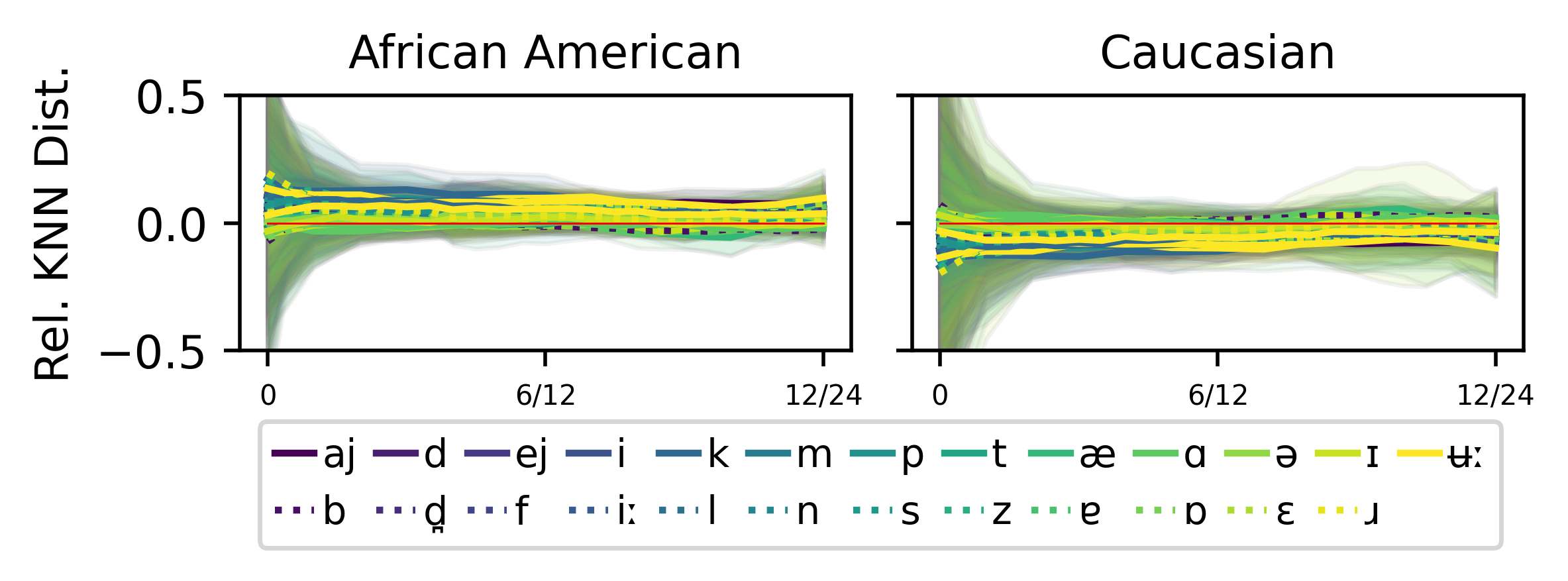}
  \caption{Relative KNN distance between embeddings of the same phoneme for the same speaker at each layer. Red line indicates SG parity.}
  \label{fig:knn_distance}
\end{figure}

Figure \ref{fig:knn_distance_absolute} shows the absolute KNN distance for all phonemes across all layers for ASR finetuned models. Note how the KNN distance decreases layer by layer over the first several model layers for every phoneme and model - this is logical, given that PR accuracy tends to increase over these layers, thus distance between embeddings should decrease \cite{toyota_layerwise_interpretability}.

Figure \ref{fig:knn_distance} shows the relative KNN distance (with respect to the average over all SGs) between closest pairs of phoneme embeddings for the same phoneme and speaker at each layer, with each ASR-finetuned S3M superimposed. As before, we note that all models exhibit similar behavior.

Furthermore, our KNN distance results bear a resemblance to those for PR in the previous section. Females and males have similar KNN distances for all phonemes (the mean for some phonemes is lower for females, though the differences are not significant due to high variance between speakers). Adults have much lower KNN distance for all phonemes than do children; likewise for Caucasians compared with African Americans. Native speakers likewise have lower KNN distance for most phonemes than Latinos and Asians by the final layers, though middle layers are more varied. 

\begin{figure}
  \centering
  \includegraphics[width=\linewidth]{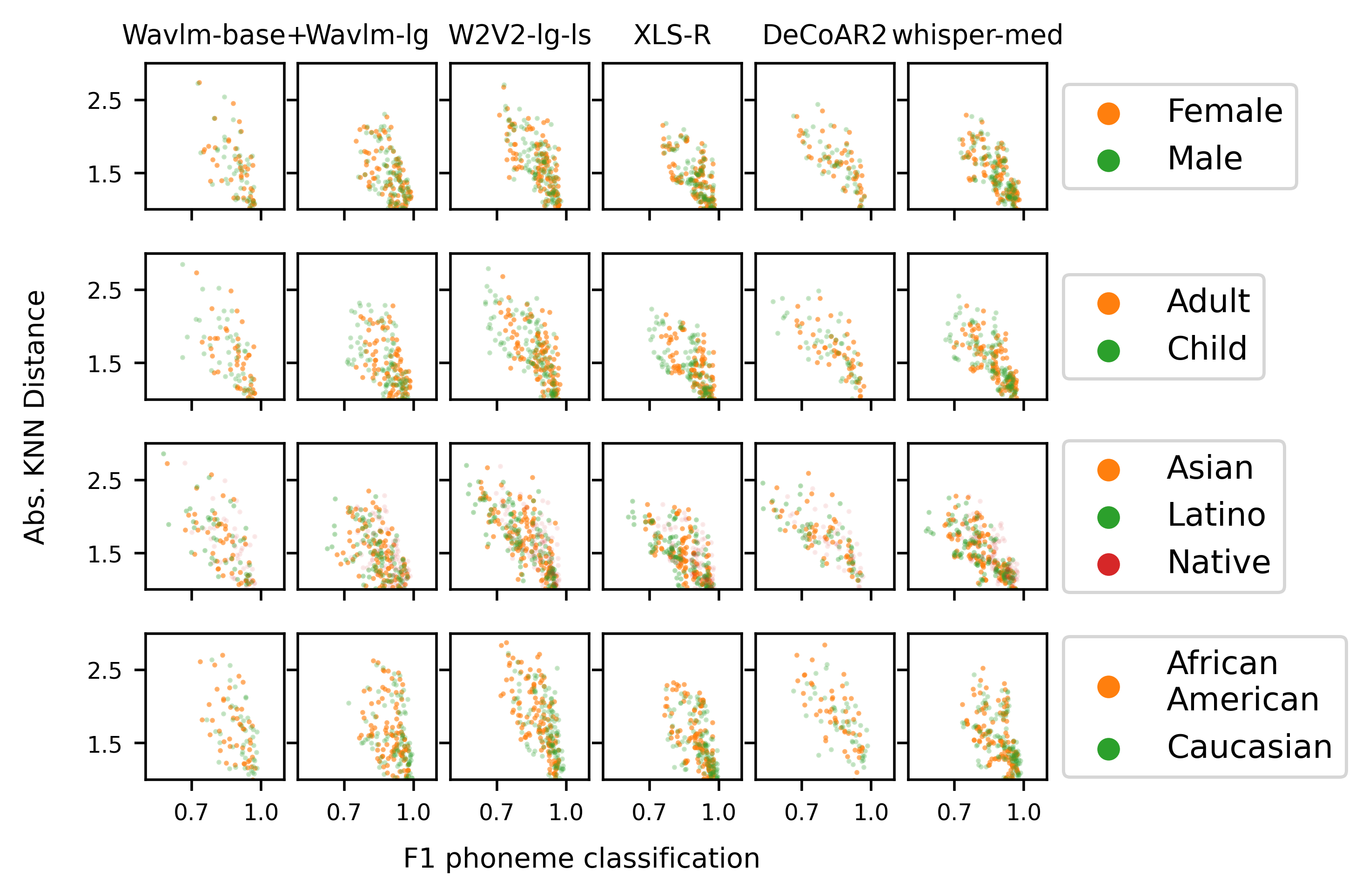}
  \caption{Relationship between KNN distance and phoneme classification rate for best performing layers at PR for each ASR-finetuned model. Each datapoint is a single phoneme at a single layer for a single SG. Statistically significant Pearson's r with $p < 0.001$ for every model and every demographic variable.}
  \label{fig:phoneme_classification_vs_knn_distance}
\end{figure}

To further establish the connection between phoneme classification and KNN distance, we examine the direct relationship between the two of them for the best performing layers in PR, separated by SG. Figure \ref{fig:phoneme_classification_vs_knn_distance} shows that phonemes that achieve higher classification accuracy also have lower KNN variance. This supports the utility of KNN distance as a metric for understanding phoneme modeling error overall, as well as for understanding fairness in PR. While we have no evidence for a causal relationship between the two, this observation does suggest that learning strategies that minimize KNN distance could in turn help improve ASR performance.

\section{Finetuning using CTC + DET/DAT}

\begin{figure}[t]
  \centering
  \includegraphics[width=\linewidth]{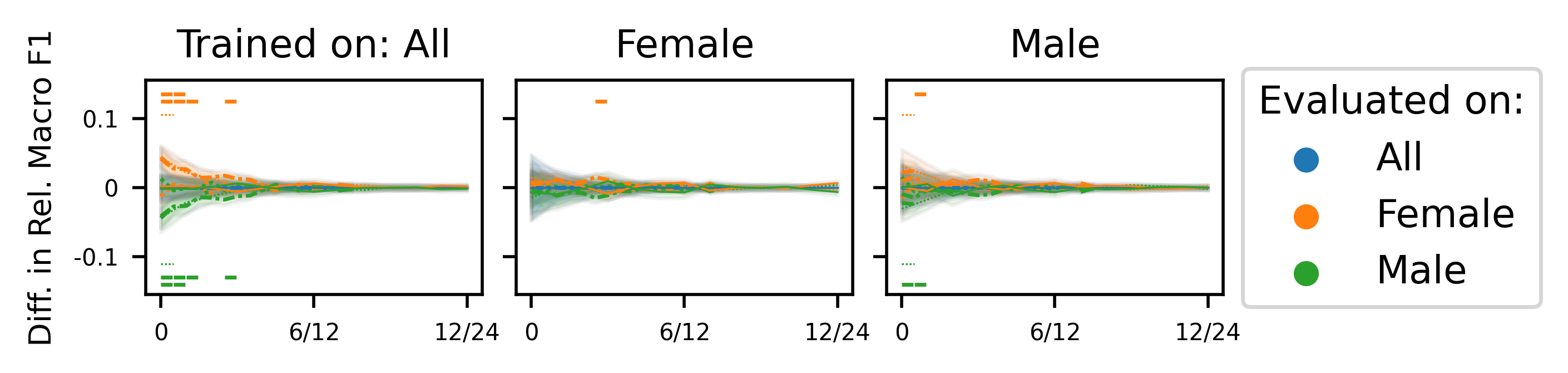}\hfill
  \includegraphics[width=\linewidth]{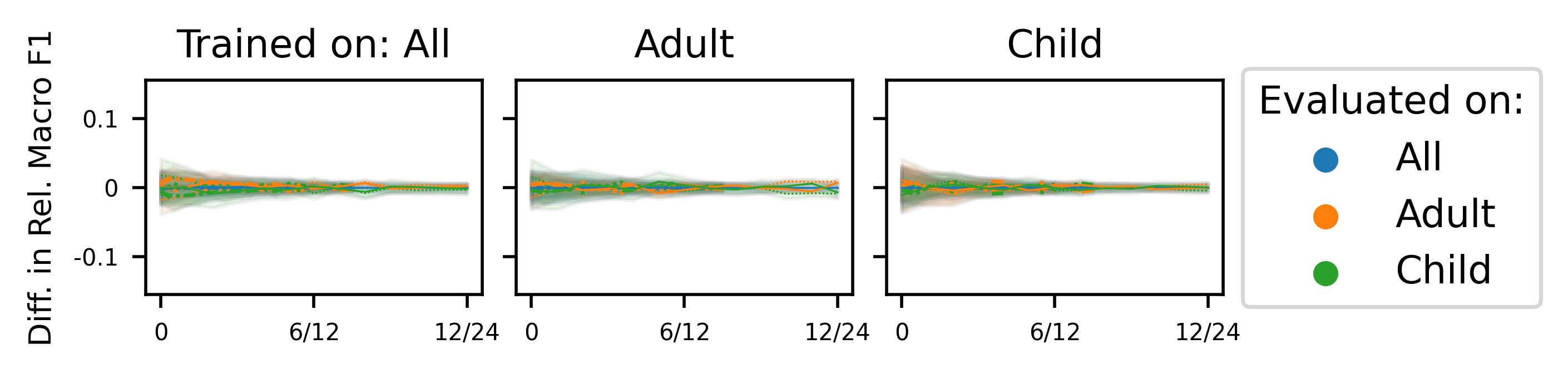}\hfill
  \includegraphics[width=\linewidth]{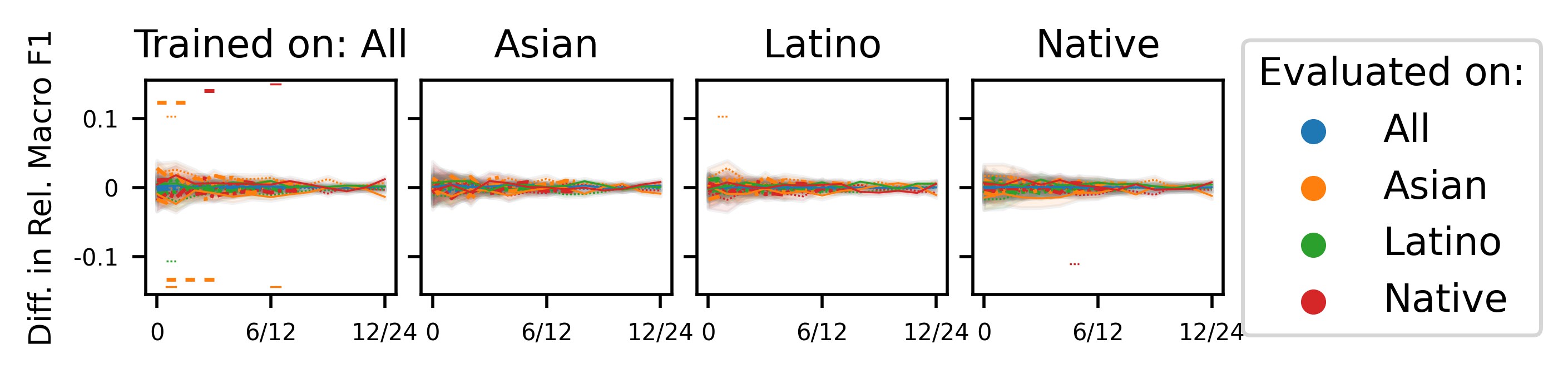}\hfill
  \includegraphics[width=\linewidth]{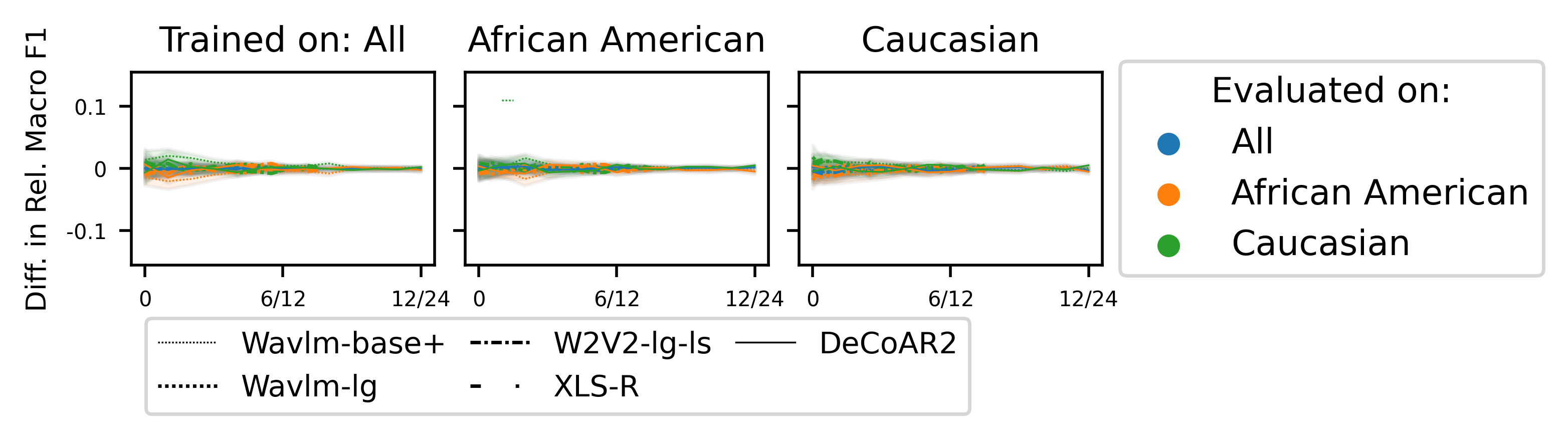}
  \caption{Difference in relative (to balanced training data) macro F1 phoneme prediction results between S3Ms finetuned on ASR and S3Ms finetuned on ASR \textit{and} DET/DAT. Values $<0$ imply that a SG performs worse relative to an average of all SGs for probes trained on CTC + DET/DAT than on just ASR. Horizontal lines on top and bottom of each figure denote statistical significance (for $p<0.05$) for a difference in relative F1 $</> 0$ respectively on 2-sided 2-sample t-test.}
  \label{fig:pheme_recognition_asr_vs_detDat}
\end{figure}

We investigate the effect that using DET/DAT during ASR finetuning has both on the effect of training PR probes on single SGs, as well as KNN distance. Figure \ref{fig:pheme_recognition_asr_vs_detDat} shows the difference between relative macro F1 scores on S3Ms finetuned on ASR alone vs CTC + DET/DAT. If DET/DAT were to reduce embedding bias, we would see curves below 0 for SGs which are disadvantaged due to embedding bias in the normal ASR setting, and above 0 for SGs which are advantaged. However, we see that there is no statistically significant difference between embeddings trained using DET/DAT for any SG in any demographic variable apart from the earliest layers where variance is high and PR is poor. This shows that if there is any embedding bias, DET/DAT is not reducing it.

\begin{figure}[t]
  \centering
  \includegraphics[width=\linewidth]{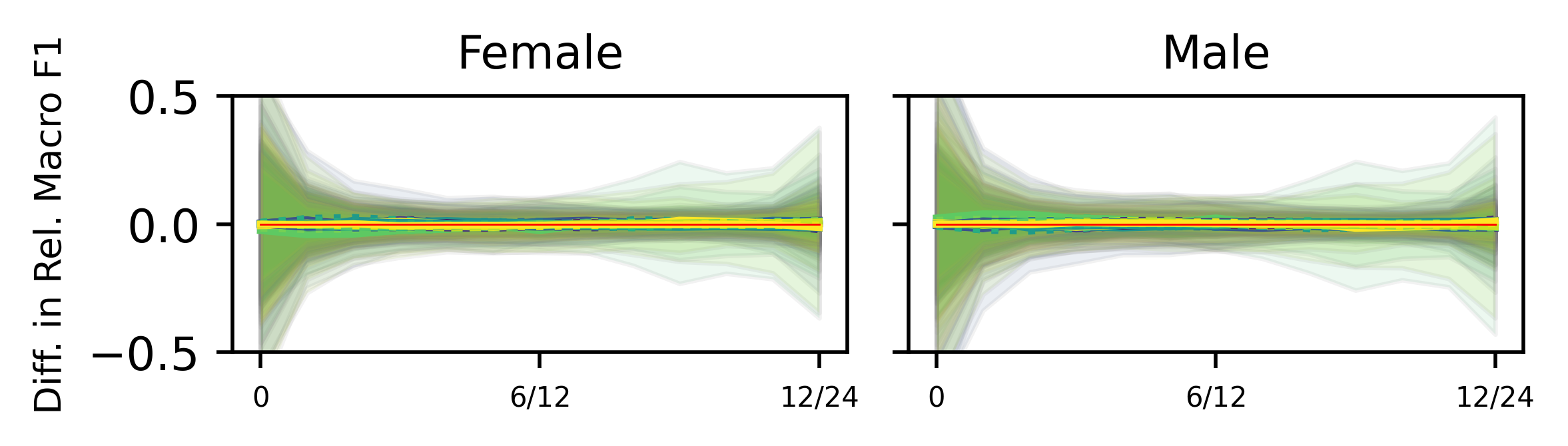}\hfill
\includegraphics[width=\linewidth]{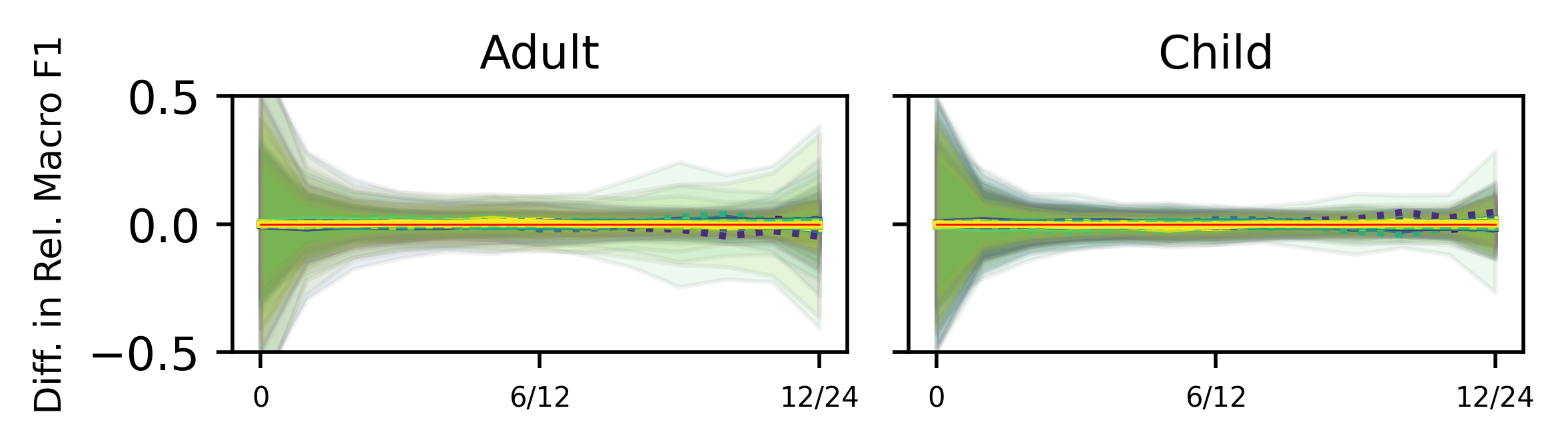}\hfill
\includegraphics[width=\linewidth]{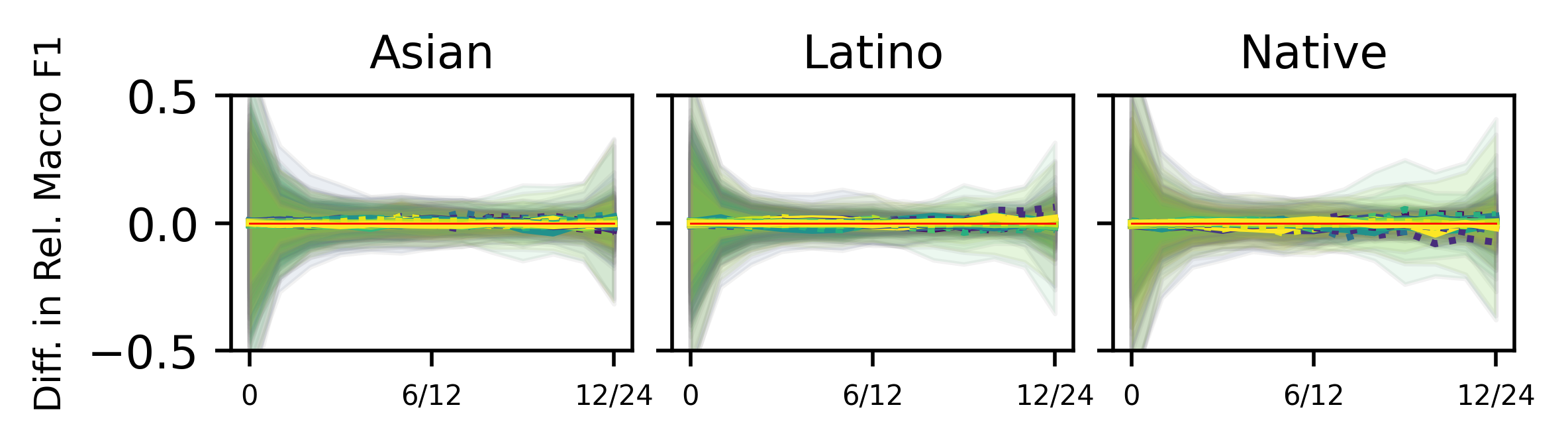}\hfill
\includegraphics[width=\linewidth]{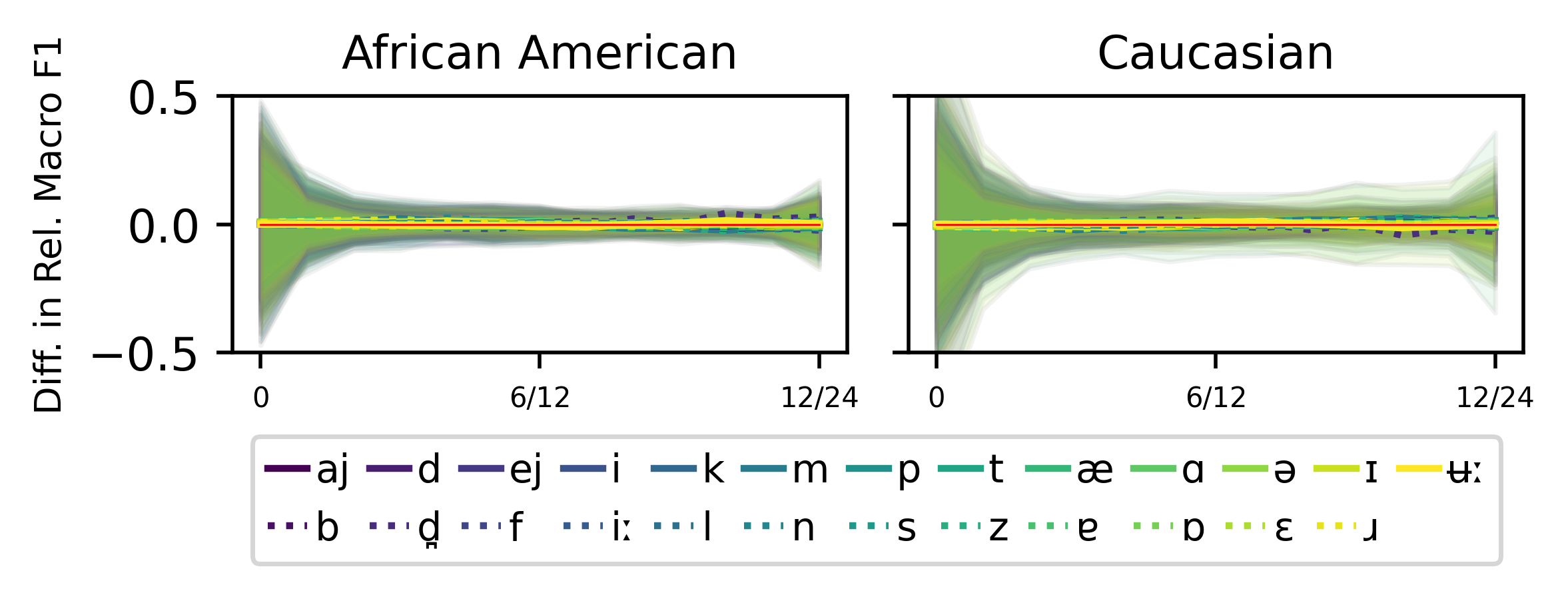}
  \caption{Difference in relative KNN distance between embeddings of the same phoneme for the same speaker at each layer. Values below zero indicate that a SG has lower average relative KNN distance for that phoneme on CTC + DET/DAT than on ASR only; however, we observe negligible deviation from zero.}
  \label{fig:knn_distance_asr_vs_detDat}
\end{figure}

Figure \ref{fig:knn_distance_asr_vs_detDat} shows the difference in relative KNN distances for models finetuned for ASR vs CTC + DET/DAT with respect to just CTC. If DET/DAT were to reduce relative KNN variance for a SG with previously higher-than-average KNN variance, we would see curves below the zero line. However, as before, we find negligible impact on KNN distance when finetuning using DET/DAT - if anything, it helps the already advantaged SGs Native speakers, adults, and Caucasians ever so slightly for some phonemes at later layers. 

These findings complement \citet{layerwise_fairness_interspeech_todo_arxiv} which shows in more abstract terms that ASR finetuning with DET/DAT has no tangible effect on fairness. Our findings provide a clearer answer as to why: we see little by way of embedding bias in ASR-finetuned S3Ms to begin with, particularly in later layers. DET/DAT is meant to enhance speaker information in middle layers and erase it in later layers; however, unfairness in phoneme prediction doesn't appear to be an embedding bias issue, and more of a variance issue. Thus, it is not totally surprising that this method is ineffective at bringing forth fairer phoneme embeddings.

\section{Discussion and outlook}

The purpose of this study is to provide tools for better understanding the differences in phoneme embeddings between speaker groups as a means to better understand unfairness in ASR. The more insight we have into how S3Ms work, as well as how they fail for specific SGs, the better we can be at making precise interventions to improve fairness.

We show that multiple different speech encoder models produce phoneme embeddings for speakers of particular SGs that can exhibit both embedding bias and/or unequal variance depending on layer. However, we show that bias is less significant in later layers, while variance remains higher for certain SGs than others. Furthermore, we show that finetuning for ASR does next to nothing to change this behavior compared to pretrained S3Ms, even when finetuning using the fairness enhancing DET/DAT. This shows the limitations of current speech encoder systems in fairly modeling phonemes from different SGs. In demonstrating that high variance is strongly associated with poor PR results, we conclude that current strategies for fairness enhancement are potentially ill-equipped to improve fairness, as they are designed to treat embedding bias more than unequal variance.

Reducing the problem of demographic fairness to a geometric modeling problem could provide a cleaner angle of tractability for future work in designing fairer ASR systems. In particular, we encourage further research into methodology that could stabilize variance/KNN distance in phoneme embeddings, both overall as well as for particularly SGs. There is ample work in other domains of self-supervised learning working on solving this type of problem, such as using Siamese networks \cite{siamese_ssl} or maximizing the effective entropy of embeddings \cite{entropy_pretraining}. Potentially methods like supervised contrastive learning could be used to reduce KNN distance for disadvantaged SGs \cite{supervised_contrastive_learning}. Indeed, some work has already used similar methodology in speech processing applications, though not with respect to fairness \cite{phoneme_contrastive}.

% This could potentially further motivate methods that explicitly  such as phoneme-level contrastive The main drawback to this is that it requires phoneme-level alignment  %: reducing SG-level extrema in either random or systematic error on specific phonemes might lead to their being more fairly classified.

We also encourage further research into error analysis from a causal perspective. Our results are a useful step towards better understanding unfairness in ASR systems but we are still lacking answers as to what exactly is responsible for either the variance or bias we observe in phoneme embeddings. With the amelioration of voice generation models, causal intervention on speech is increasingly viable \cite{masson_phoneme_similarity_tts,vc_article,hend_gender_emnlp}. Future work could thus make precise manipulations to an input signal corresponding to demographic characteristics associated with any protected SG, to discern where and how along the layerwise computation phoneme representations begin to shift in response.

% Finally, we encourage further work into the impact of existing fairer speech processing techniques based on our error framework. This would be useful in studying why certain fairness techniques work well (or otherwise) beyond simple analyzing the resulting word error rate.

\clearpage
\section*{Limitations}

We conducted our analysis on the \sonos{} dataset primarily due to its precisely labeled and diverse metadata, allowing for fine-grained fairness analysis. However, this limits its generalizability to speech processing in general due to the controlled nature of recordings. Speech corpora with high quality annotations are few and far between - we had previously worked with Fair-speech \cite{meta_fair} but it lacks speaker IDs, so we cannot control for the number of speakers per SG, rendering our analyses much less useful.

Furthermore, we used 100- and 300-million parameter models, much smaller than those used to achieve state of the art ASR performance, such as \whisper{}-large. It is possible, due to increased capabilities of larger models, that the effects we note in our study would be different on such models.

Finally, our analyses are based on automatically extracted phoneme alignments. This is a potential source for experimental noise due to alignment error in general; it also has the potential to have an outsize impact on certain SGs, resulting in noisier phoneme alignments for those SGs. Thus, unfair phoneme alignment has the potential to further compound error and it is unclear how much of the KNN distance is attributable to this effect.

\section*{Acknowledgements}
This research has been funded by the French National Research Agency (ANR), project "E-SSL" (ANR-22-CE23-0013). It was also partially supported by ANR through the MIAI "AI \& Language" chair and the MIAI "Socialization and Language at School" chair (ANR-23-IACL-0006). This work was performed using HPC resources from GENCI at IDRIS under the allocations 2023-A0151014633, 2024-A0171014633 on the Jean Zay supercomputers.

% \subsection{Analysis of phoneme classifier logit entropy}

% A further measure of error typology in phoneme embedding comes from measuring the entropy of the output space of the phoneme classifiers trained in Section \ref{sec:indirect_explanation}. For incorrectly classified phonemes, if the entropy is high (e.g. high probability for many phonemes 

\FloatBarrier
\newpage
\bibliography{these,underReview} %these

\clearpage
\appendix

\section{Motivation for KNN distance metric}
\label{sec:knn_motivation}

\begin{figure*}[t]
  \centering
    \includegraphics[width=\linewidth]{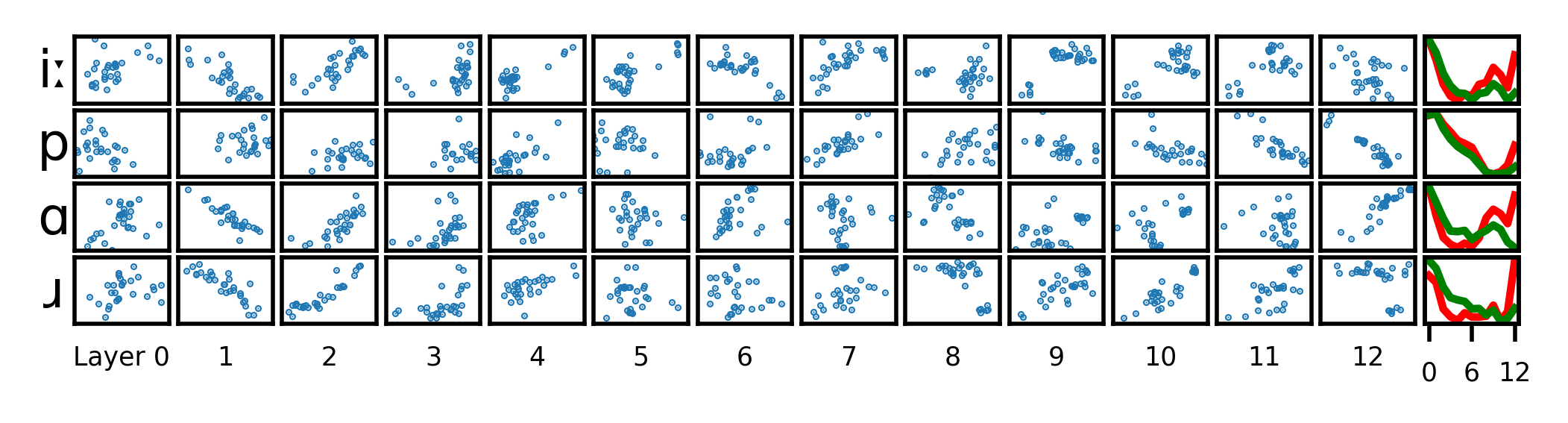}
  \caption{2-dimensional PCA decompositions for phoneme embeddings for speaker 440 for four phonemes for \wavlm{}-large finetuned on ASR, for each layer. Final column plots the average distance from the mean (red) and KNN distance (green).}
  \label{fig:kmeans_vs_variance}
\end{figure*}

Our first intuition when considering calculating random error in phoneme embeddings was to simply calculate the Euclidean distance from each phoneme embedding to the mean for each phoneme and speaker, i.e. the variance. However, we realized that was a potentially foolish proposition, as it assumes a distribution with a single mode or mean. There is no reason to assume this to be the case for phoneme embeddings, and indeed, Figure \ref{fig:kmeans_vs_variance} provides a visualization showing it to be otherwise for some phonemes. At each layer, we train a 2-dimensional PCA reduction on embeddings for all phonemes, then visualize several phonemes individually. Note that some layers' embeddings for some phonemes are clearly bimodal starting in later layers (e.g. `r' in particular), while none is a 2-dimensional Gaussian, apart from the earliest layers. Indeed, this is a mere 2-dimensional reproduction; there may be further multidimensional means that are tamped down by PCA.

With that in mind, Figure \ref{fig:kmeans_vs_variance} provides a motivating example for why KNN-distance is a more suitable metric for measuring phoneme distribution variance. It remains low for a multimodal distribution where each mode has low variance. The red curve (Euclidean distance from the mean) jumps up in later layers, despite the phonemes being being increasingly well defined; on the other hand, the green curve (KNN distance) often decreases monotonically and certainly doesn't jump when multimodality appears.

% \clearpage
\section{Extraction of phoneme embeddings}
\label{sec:embedding_extraction}

We use the Montreal Force Aligner \cite{MFA} version 3.3.4 to obtain phoneme-level alignments for the entire \sonos{} corpus, which we use to extract phoneme embeddings from each S3M at every layer. We use the "english\_mfa'' model and corresponding ``english\_us\_mfa'' alignment dictionary, due to our corpus being focused on American English. We select the 25 phonemes which are most frequently voiced over the corpus to include as full a phoneme embedding dataset as possible (see Figure \ref{fig:by_phoneme_analysis_probe_training} for a list of phonemes). It was unfortunately impractical to select all phonemes included in ``english\_mfa'', as many speakers didn't use some rarer phonemes enough times for them to be included in phoneme-level analysis.

Using the phoneme alignments, we then extract phoneme alignments at every layer of each S3M. Following \citet{mean_pooling_sid}, we first normalize all frame embeddings at each layer over the entire utterance to remove global utterance-level information. Failing to remove utterance-level information would potentially add distracting information that could add noise to KNN distance calculations (although \cite{orthogonality_first} shows that phoneme and speaker information are modeled in orthogonal subspaces, thus in theory this should be redundant). After normalizing, we follow \citet{toyota_layerwise_interpretability} to construct our phoneme embeddings by mean-pooling over frames corresponding to a given phoneme. However, rather than taking all frames that fall within the given window, we take only the middle third of frames to reduce co-articulation effects. For each speaker, we sample 30 instances of each of the 25 most common phonemes at each transformer layer. We then filter out obvious alignment errors on a by-phoneme, by-speaker basis by discarding phoneme embeddings that are over three z-scores away from the mean over all instances of each phoneme/speaker.

The first row of Table \ref{tab:phoneme_rec} shows the overall rates of PR for the best layers each finetuned S3M. That we are able to achieve such a high rate of PR validates our embedding extraction algorithm.

% \clearpage
\section{Analysis of pretrained S3Ms}
\label{sec:pretrained_PR}

\begin{figure}[h!]
  \centering
  \includegraphics[width=\linewidth]{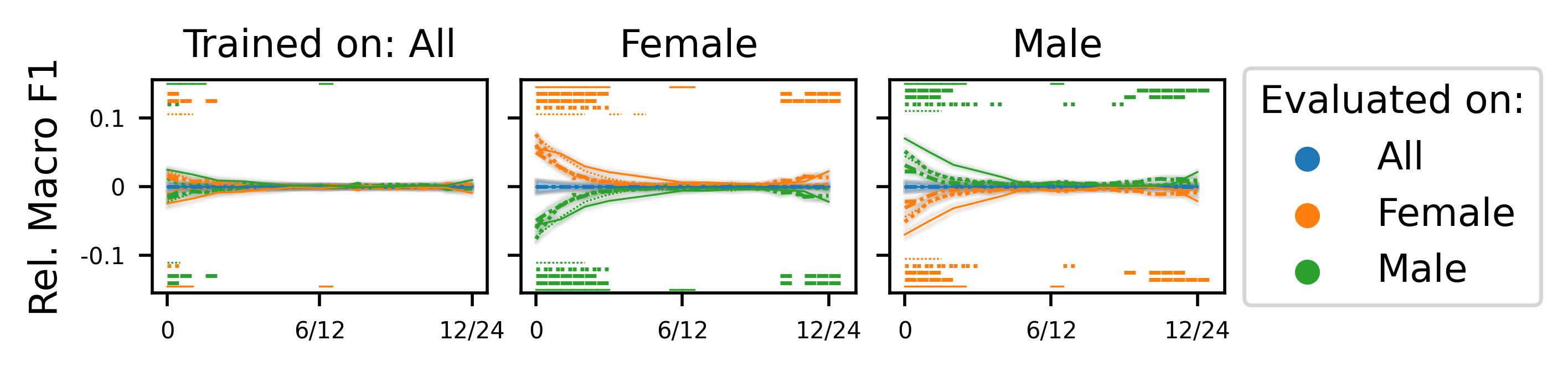}\hfill
  \includegraphics[width=\linewidth]{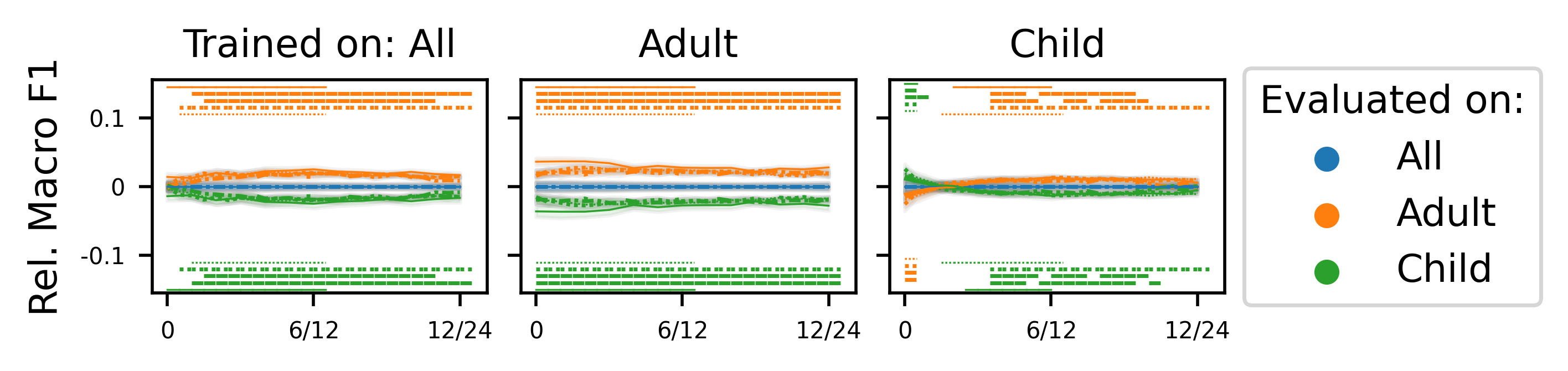}\hfill
  \includegraphics[width=\linewidth]{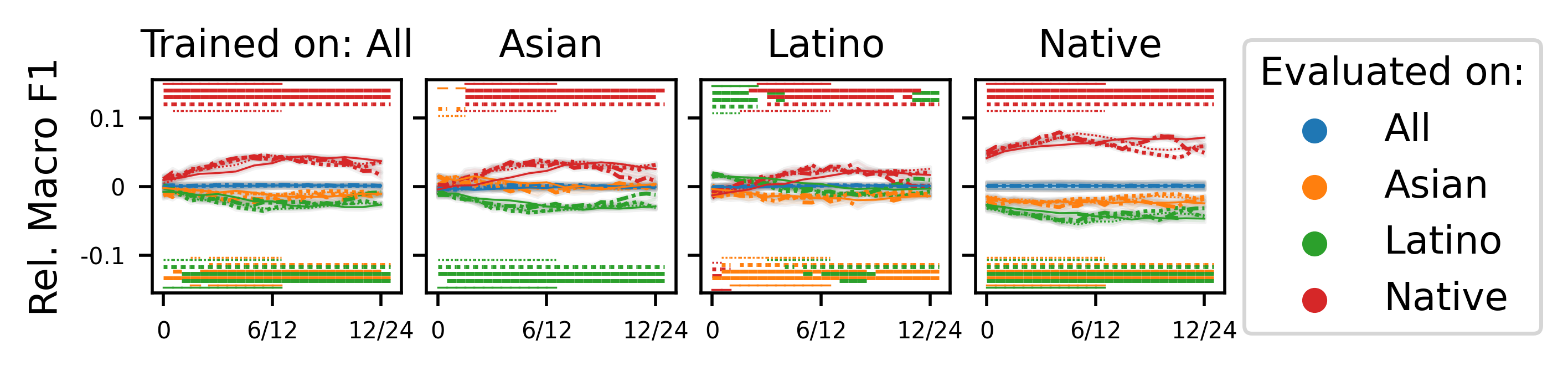}\hfill
  \includegraphics[width=\linewidth]{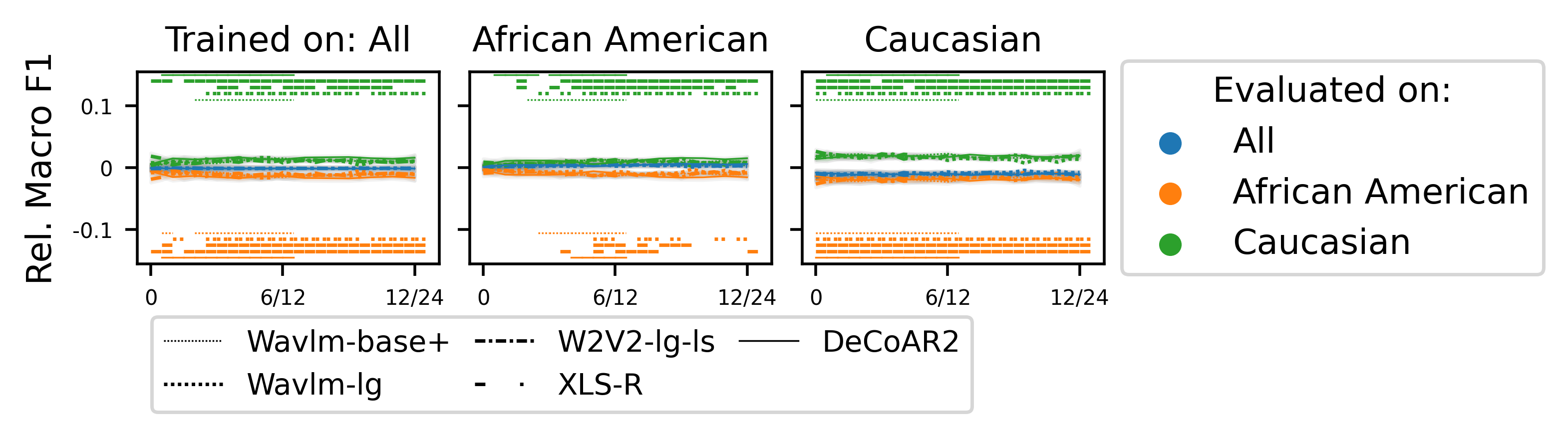}
  \caption{Macro F1 phoneme classification scores, \textbf{relative to the macro average over all SGs} (e.g. men and women) for the corresponding demographic variable (e.g. gender) for \textbf{pretrained S3Ms}. Values $>0$ indicate that SG has a  better-than-average macro F1 (e.g. top left: when trained on all data, females have above-average macro F1 performance in layer 0). Horizontal lines on top and bottom of each figure denote statistical significance (for $p<0.05$) for relative F1 $</> 0$ respectively on 1-sided 1-sample t-test.}
  \label{fig:by_training_pretrain}
\end{figure}

\begin{figure}[h!]
  \centering
  \includegraphics[width=\linewidth]{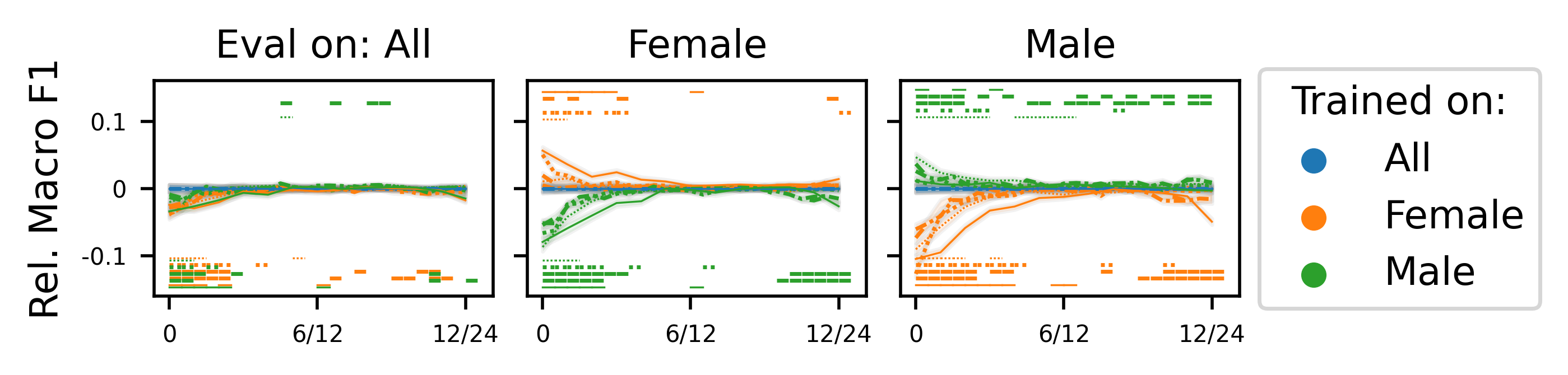}\hfill
  \includegraphics[width=\linewidth]{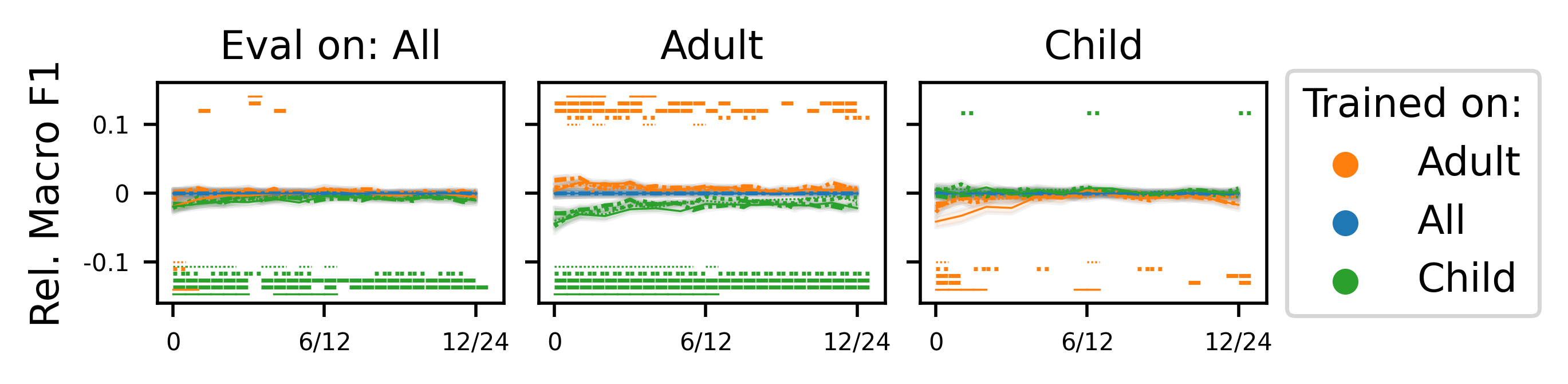}\hfill
  \includegraphics[width=\linewidth]{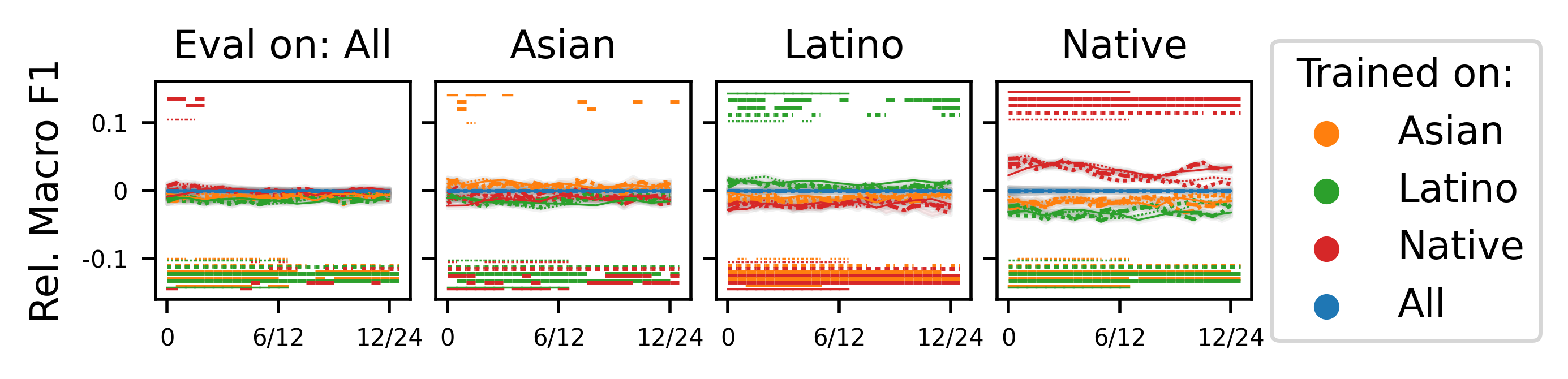}\hfill
  \includegraphics[width=\linewidth]{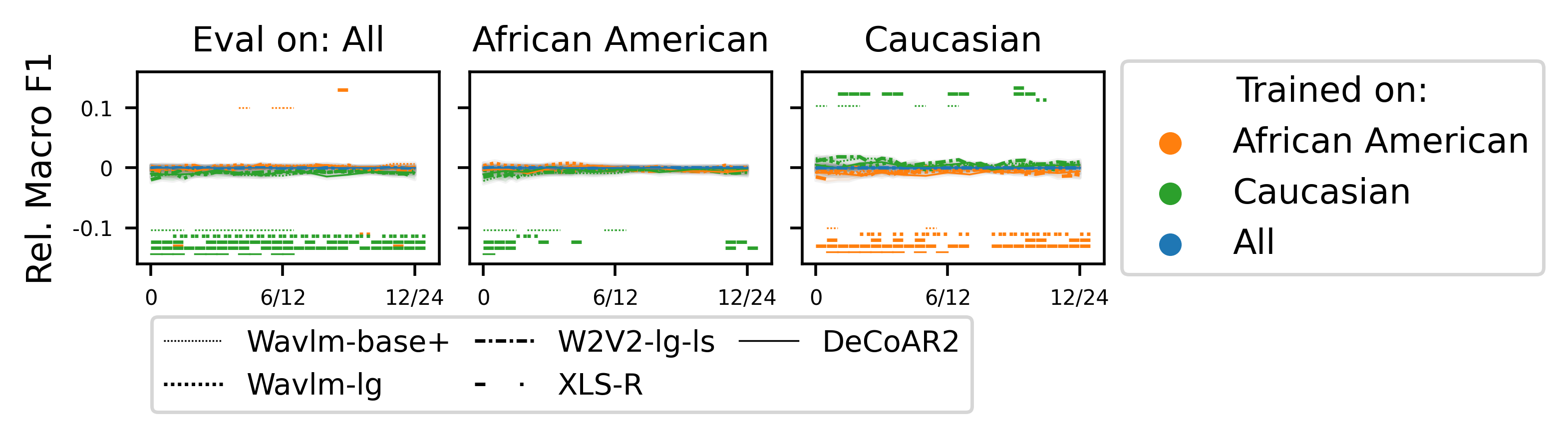}
  \caption{Macro F1 phoneme classification scores, \textbf{relative to probe training on a balanced dataset} for SGs from the corresponding demographic variable, for \textbf{pretrained S3Ms}. Values $>0$ indicate that training on that SG results in better performance than for a balanced dataset (e.g. top left: when phoneme classifiers are trained on only men, men have a higher PR than for probe training on balanced data). Horizontal lines on top and bottom of each figure denote statistical significance (for $p<0.05$) for relative F1 $</> 0$ respectively on 1-sided 1-sample t-test.}
  \label{fig:by_eval_pretrain}
\end{figure}

\begin{figure}[t]
  \centering
  \includegraphics[width=\linewidth]{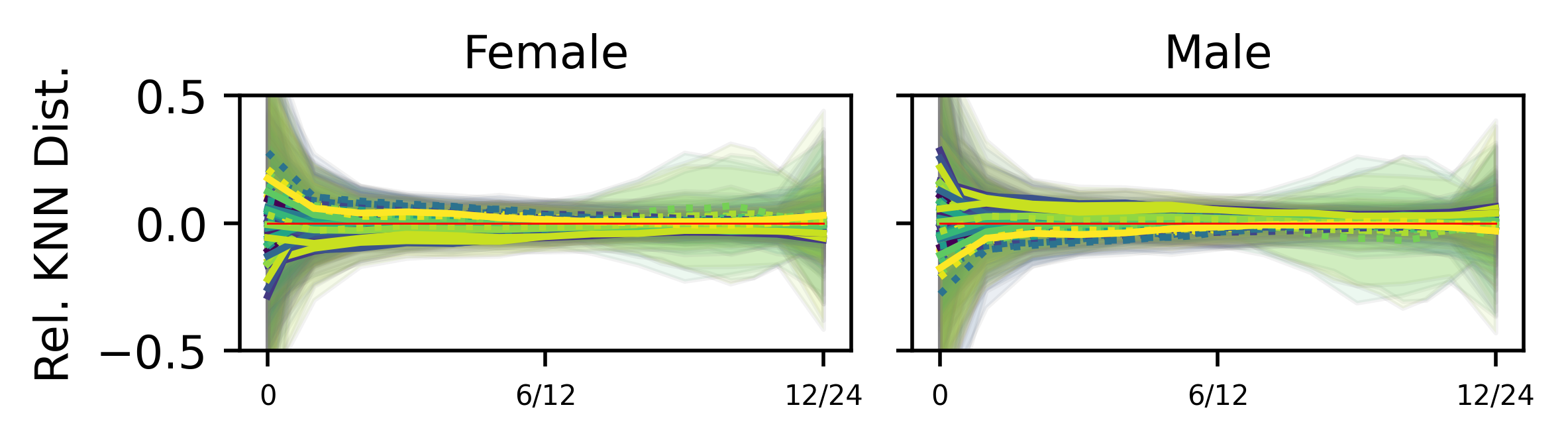}\hfill
\includegraphics[width=\linewidth]{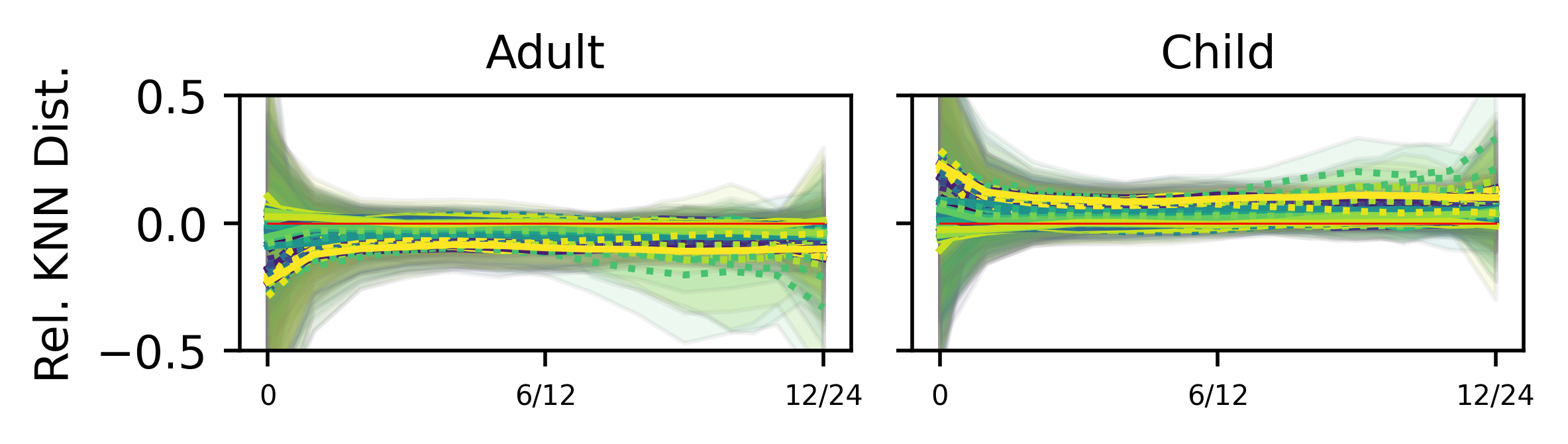}\hfill
\includegraphics[width=\linewidth]{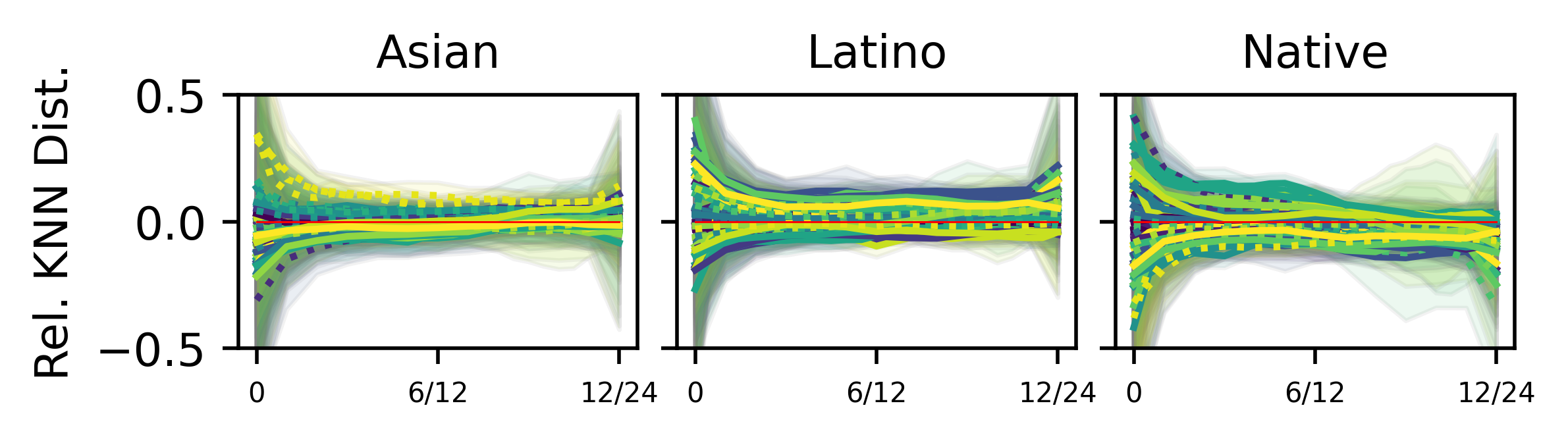}\hfill
\includegraphics[width=\linewidth]{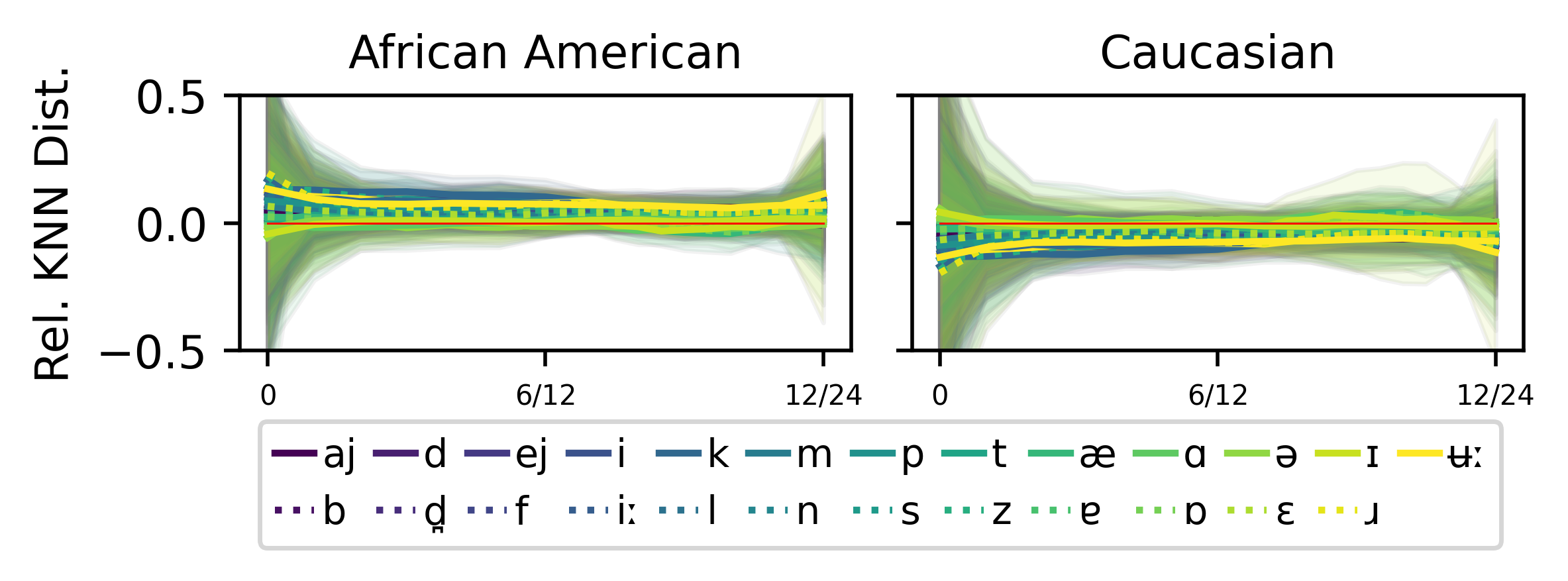}
  \caption{Relative KNN distance between embeddings of the same phoneme for the same speaker at each layer of \textbf{pretrained} S3Ms.}
  \label{fig:knn_distance_pretrained}
\end{figure}

We replicate PR experiments on pretrained S3Ms. Figures \ref{fig:by_training_pretrain} and \ref{fig:by_eval_pretrain} are analogous to Figures \ref{fig:by_training_finetune} and \ref{fig:by_eval_finetune}. We note the same patterns in pretrained models as their ASR-finetuned complements. We likewise repeat out KNN distance analyses on pretrained S3Ms in Figure \ref{fig:knn_distance_pretrained}. (We excluded \wtvt{} models to avoid visual contamination by their strange behaving final several layers \cite{toyota_layerwise_interpretability}). Note that pretrained models exhibit the same patterns of increased KNN variance in the same SGs as ASR finetuned models in Figure \ref{fig:knn_distance}.

\cleardoublepage
\section{By-phoneme analysis of in-domain probe training}
\label{sec:by_phoneme_analysis_probe_training}

\begin{figure*}
  \centering
    \includegraphics[width=\linewidth]{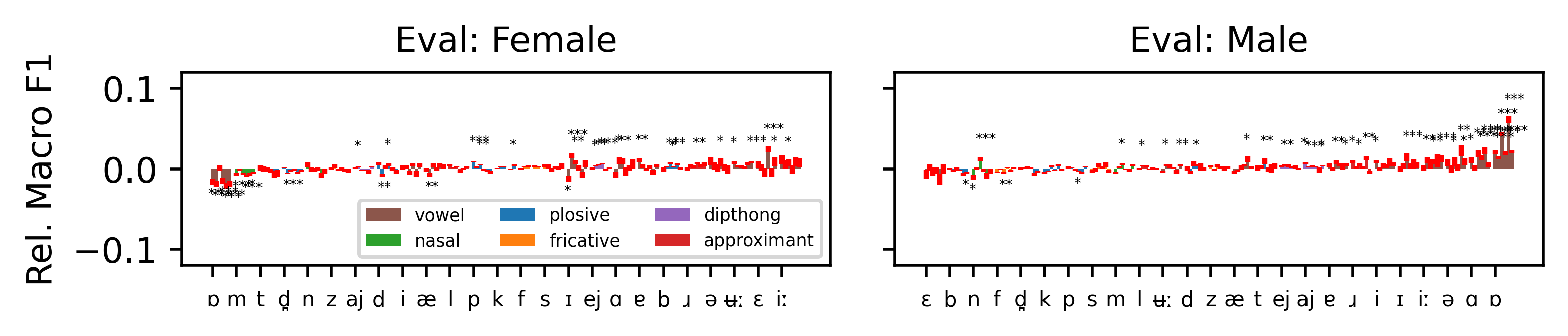}\hfill
    \includegraphics[width=\linewidth]{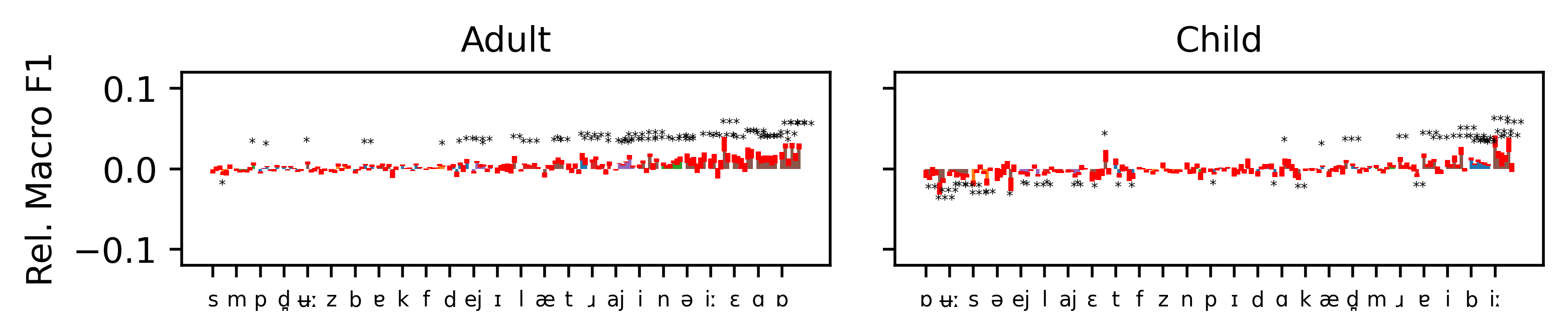}\hfill
    \includegraphics[width=\linewidth]{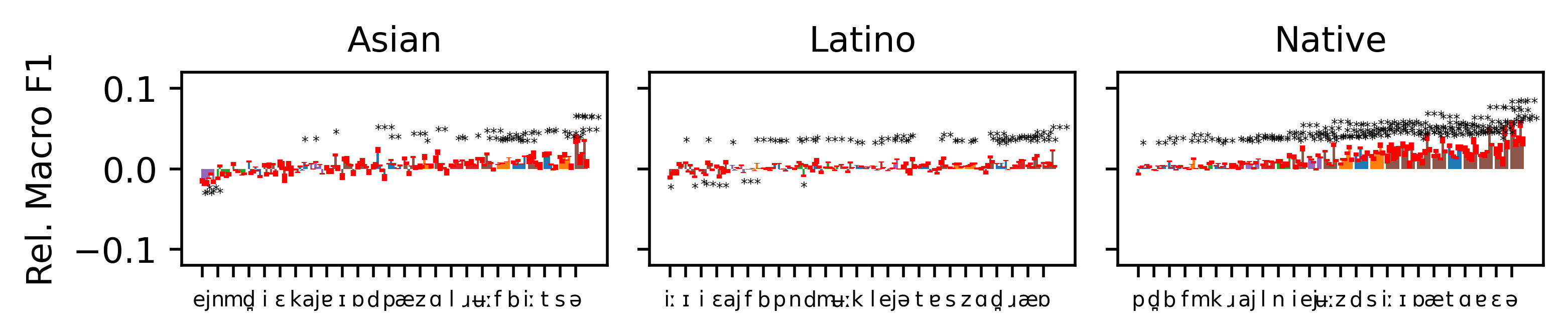}\hfill
    \includegraphics[width=\linewidth]{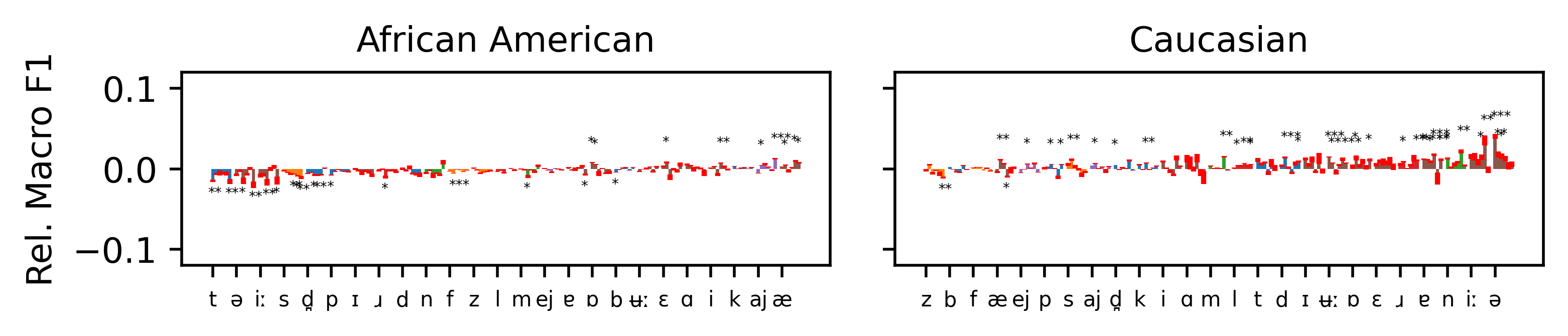}
  \caption{Macro F1 phoneme classification scores, \textbf{relative to probe training on a balanced dataset} for SGs from the corresponding demographic variable, for each phoneme separately, aggregated over the best-performing layers of all six encoder models. Values $>0$ indicate that training on that SG results in better performance than for a balanced dataset. Each phoneme has six bars, one per encoder model. * represents statistical significance on a 1-sample 1-sided t-test: * indicates $p < 0.05$, ** for $p < 0.01$, and *** for $p < 0.001$.}
  \label{fig:by_phoneme_analysis_probe_training}
\end{figure*}

To get a more precise picture of embedding bias, we examine which phonemes experience the biggest boost from in-domain probe training. Figure \ref{fig:by_phoneme_analysis_probe_training} shows that some phonemes receive are classified better on a phoneme classifier trained in-domain, while others experience the opposite effect. Phonemes which experience improvement are likely those which are embedded in a biased manner by S3Ms, while those which experience degradation are likely casualties of higher variance for that SG. Note that the phonemes with significant performance improvement (or impairment) for in-domain training are largely shared across encoder models, indicating that all models exhibit similar trends of systematic and random error.

% \clearpage
\section{Systematic error as linear SG separability}

We would be interested, if possible, in directly quantifying systematic error as we did random error by KNN distance. That is to say, to calculate the extent to which phonemes are modeled around separate modes for different SGs (Figure \ref{fig:error_typology} lower left). One potential method to this end is to train a linear classifier to predict SG (i.e. men or women) for every demographic variable (i.e. gender), based on phoneme embeddings. However, for this analysis to be sound, we would need filter out all utterance-level/speaker-level information from each phoneme embedding so that the classifier cannot use it in discriminating between SGs. One strategy to this end is to normalize out utterance-level information (recall Section \ref{sec:embedding_extraction}), which we have already done in data preparation.

\begin{figure}
  \centering
  \includegraphics[width=\linewidth]{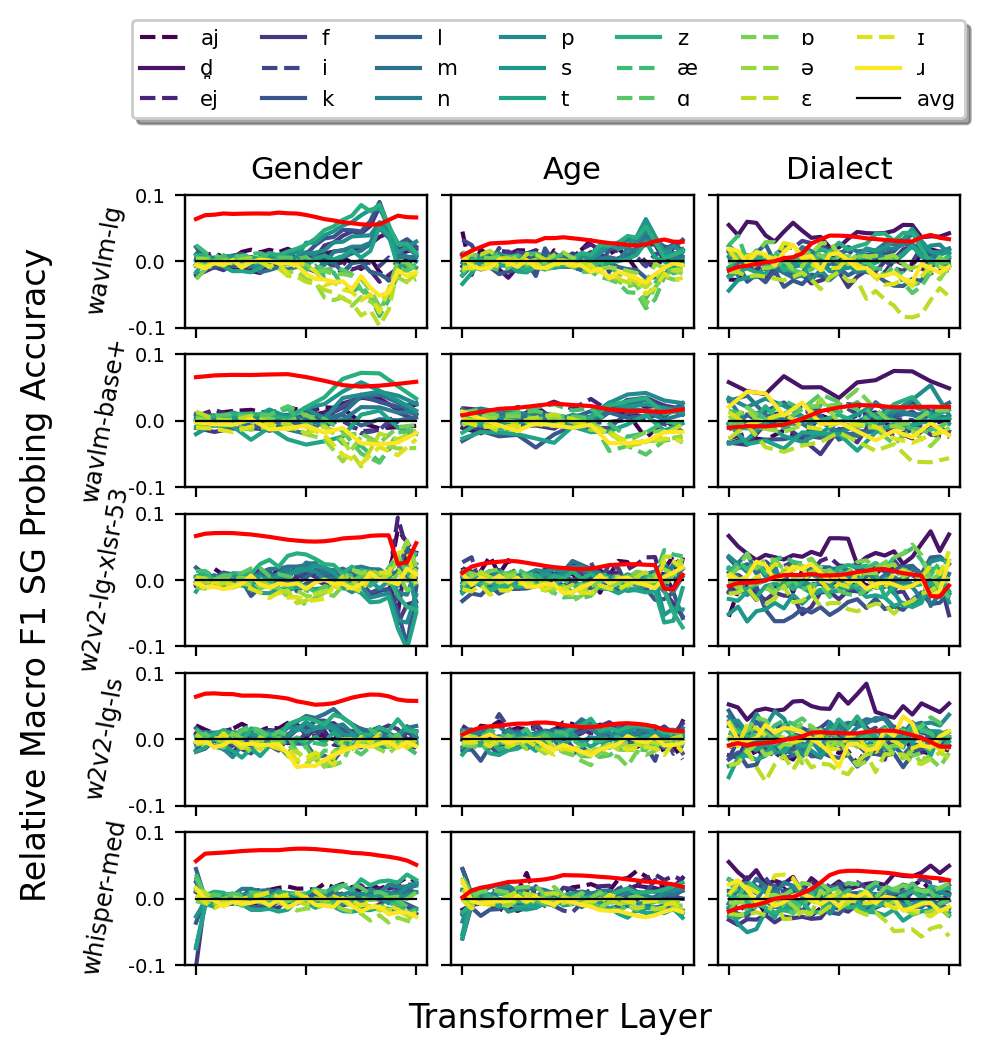}
  \caption{Relative macro F1 probing accuracy for each demographic variable (DV) based on embeddings from various phonemes. Red line is absolute average F1 over all phonemes. Scores $<0$ means less SG information is present for any given phoneme/layer than the average over all phonemes at that layer.}
  \label{fig:sgi_classify}
\end{figure}

However, in practice, we are skeptical of the completeness of this approach. Figure \ref{fig:sgi_classify} shows that we can predict each demographic variable fairly well at every layer using every single phoneme embedding. This strongly implies that global speaker-level information is retained despite normalizing over the utterance. Thus, we refer to stick with the in-domain probe training approach, which is guaranteed to take into account only SG-specific embedding bias that pertains to individual phoneme embeddings. However, future work could try and adapt methods such as linear adversarial concept erasure to ensure global speaker information is excluded \cite{linear_adversarial_concept_erasure}.

% However, as \citet{orthogonality_first} points out, speaker and phoneme information are modeled in orthogonal subspaces of phoneme embeddings, so it is possible that SG information is automatically discarded by phoneme classifiers as noise. That said, we again use SG classification performance as a comparative metric between phonemes, therefore utterance-level information will likely be smoothed out when comparing between phonemes.

\end{document}